\documentclass[acmtog, authorversion,nonacm]{acmart}

\usepackage{booktabs} 

\newcommand{\figref}[1]{Figure~\ref{#1}}
\newcommand{\tabref}[1]{Table~\ref{#1}}
\newcommand{\secref}[1]{Section~\ref{#1}}

\definecolor{mytbcol}{RGB}{175,227,246}
\usepackage{colortbl}
\usepackage{xcolor}
\usepackage{multirow}
\usepackage{array}
\usepackage[misc]{ifsym}

\usepackage[ruled]{algorithm2e} 

\SetAlFnt{\small}
\SetAlCapFnt{\small}
\SetAlCapNameFnt{\small}
\SetAlCapHSkip{0pt}

\begin{document}
\title{OmniPart: Part-Aware 3D Generation with Semantic Decoupling and Structural Cohesion}

\author{Yunhan Yang}
\authornote{Equal contribution.}
\affiliation{%
 \institution{The University of Hong Kong}
 \country{China}}
\email{yhyang.myron@gmail.com}

\author{Yufan Zhou}
\authornotemark[1]
\affiliation{%
 \institution{Harbin Institute of Technology}
 \country{China}}
\email{yfzhou@stu.hit.edu.cn}

\author{Yuan-Chen Guo}
\affiliation{%
 \institution{VAST}
 \country{China}}
\email{imbennyguo@gmail.com}

\author{Zi-Xin Zou}
\affiliation{%
 \institution{VAST}
 \country{China}}
\email{zouzx1997@gmail.com}

\author{Yukun Huang}
\affiliation{%
 \institution{The University of Hong Kong}
 \country{China}}
\email{kunh6414@gmail.com}

\author{Ying-Tian Liu}
\affiliation{%
 \institution{VAST}
 \country{China}}
\email{liuyingt23@mails.tsinghua.edu.cn}

\author{Hao Xu}
\affiliation{%
 \institution{Zhejiang University}
 \country{China}}
\email{haoxu38@outlook.com}

\author{Ding Liang}
\affiliation{%
 \institution{VAST}
 \country{China}}
\email{liangding1990@163.com}

\author{Yan-Pei Cao}
\authornote{Corresponding author.}
\affiliation{%
 \institution{VAST}
 \country{China}}
\email{caoyanpei@gmail.com}

\author{Xihui Liu}
\authornotemark[2]
\affiliation{%
 \institution{The University of Hong Kong}
 \country{China}}
\email{xihuiliu@eee.hku.hk}

\begin{abstract}
The creation of 3D assets with explicit, editable part structures is crucial for advancing interactive applications, yet most generative methods produce only monolithic shapes, limiting their utility.
We introduce OmniPart, a novel framework for part-aware 3D object generation designed to achieve high semantic decoupling among components while maintaining robust structural cohesion. 
OmniPart uniquely decouples this complex task into two synergistic stages: (1) an autoregressive structure planning module generates a controllable, variable-length sequence of 3D part bounding boxes, critically guided by flexible 2D part masks that allow for intuitive control over part decomposition without requiring direct correspondences or semantic labels; and (2) a spatially-conditioned rectified flow model, efficiently adapted from a pre-trained holistic 3D generator, synthesizes all 3D parts simultaneously and consistently within the planned layout. Our approach supports user-defined part granularity, precise localization, and enables diverse downstream applications.
Extensive experiments demonstrate that OmniPart achieves state-of-the-art performance, paving the way for more interpretable, editable, and versatile 3D content.
Project page: \href{https://omnipart.github.io/}{https://omnipart.github.io/}.
\end{abstract}


\begin{teaserfigure}
  \includegraphics[width=\textwidth]{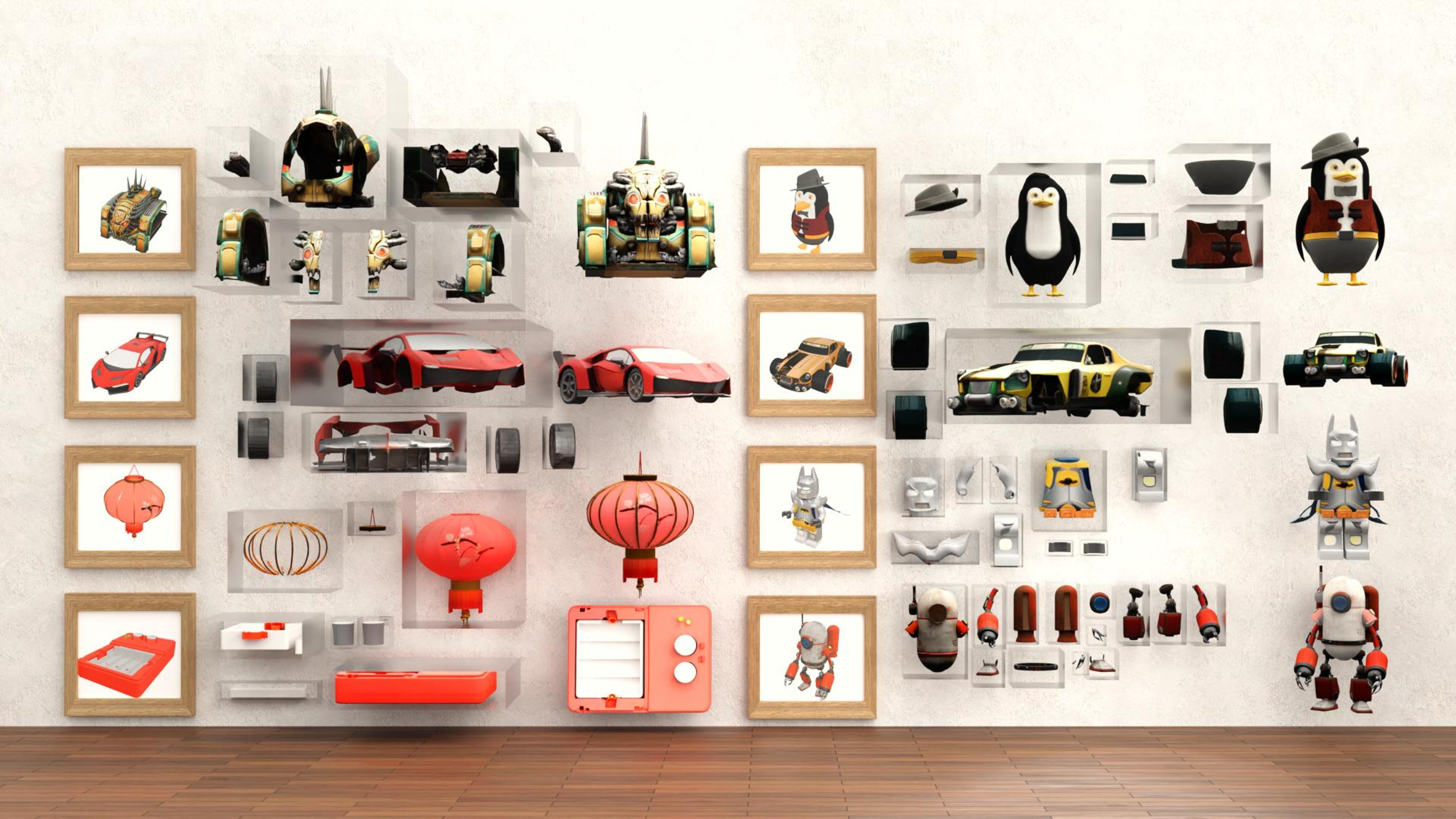}
  \caption{
  \textbf{OmniPart: Generating Complex 3D Objects as Compositions of Controllable Parts.} From simple 2D images (shown framed) and mask inputs, OmniPart first plans a 3D part structure using an autoregressive model, then synthesizes all high-quality, textured parts simultaneously (individual components within transparent displays). These seamlessly merge into a coherent object, offering explicit part control for enhanced editing, customization, and animation.
}
  \label{fig:teaser}
\end{teaserfigure}

\maketitle

\section{Introduction}
The creation of rich, interactive 3D worlds is fundamental to modern visual computing, driving applications from immersive gaming and virtual reality to digital twins and robotic interaction. A persistent challenge is the generation of 3D assets that are not merely static, monolithic forms. While recent generative models produce impressive holistic 3D shapes~\cite{poole2022dreamfusion, zhang20233dshape2vecset,zhang2024clay,li2025triposg,xiang2024structured}, these often lack the \textbf{\textit{intrinsic part-based structure}} inherent to real-world objects. This structural opaqueness limits their direct use for essential tasks such as compositional editing, procedural animation, material assignment, and semantic understanding—capabilities crucial for artists, developers, and downstream systems.

Part-aware 3D generation~\cite{chen2024partgen,yang2025holopart,liu2024part123}, which explicitly models objects as assemblies of semantically meaningful components, offers a path to overcome these limitations. However, creating such structured assets requires a careful balance: achieving \textit{\textbf{low semantic coupling}} (parts are distinct and independently addressable) while ensuring \textbf{\textit{high structural cohesion}} (parts form a plausible, integrated whole). Prior work has often struggled here. Approaches reconstructing from 2D part segmentations~\cite{liu2024part123,chen2024partgen} suffer from geometric inconsistencies and insufficient detail fidelity, primarily due to challenges in view aggregation and the inherent limitations of 2D models in capturing complete 3D part geometry. Conversely, existing methods for direct 3D part generation~\cite{li2024pasta} may lack robust, fine-grained control over part decomposition or rely on extensive, scarce 3D part-level annotations, hindering their scalability and practical use.

We emphasize that robust and versatile part-aware 3D generation hinges on \textit{a principled decoupling of high-level structural planning from detailed part synthesis, unified by strong conditioning mechanisms}. We introduce \textbf{\textit{OmniPart}}, a novel framework that implements this principle through a synergistic two-stage pipeline. 

Initially, a \textbf{\textit{Controllable Structure Planning}} module addresses the crucial task of defining the part-level spatial layout. It autoregressively generates a variable-length sequence of 3D bounding boxes, each representing a distinct part's 3D location. A key challenge here is compositional ambiguity (e.g., limbs vs. a composite torso), which can lead to unpredictability. OmniPart resolves this by conditioning the planning on intuitive, \textit{flexible 2D part masks} (obtainable from user input or 2D models like SAM~\cite{ravi2024sam}). These masks, which delineate desired regions for part decomposition \textit{without imposing strict one-to-one correspondences or requiring explicit semantic labels}, guide the transformer-based autoregressive model to produce 3D bounding boxes reflecting the intended part structure and accommodate varying part counts. To further enhance the accuracy of this structural plan, we introduce a novel \textit{Part Coverage Loss}, ensuring that each predicted bounding box comprehensively encloses its corresponding object part.

Subsequently, building on this planned structure, a \textbf{\textit{Spatially-Conditioned Part Synthesis}} module efficiently generates consistent and high-quality 3D representations for all object parts. Given the scarcity of extensive part-level 3D annotations, our approach effectively adapts a powerful, pre-trained holistic 3D generator (TRELLIS~\cite{xiang2024structured}) based on rectified flow into a part-aware 3D generation model. The predicted bounding boxes from the first stage define spatial regions within TRELLIS's voxel representation, serving as initializations for part-wise latent codes. To promote semantic awareness and overall structural coherence, each part's voxel representation is considered alongside the full-object context, and distinctive part-aware embeddings are injected. All part latents are then jointly refined via a denoising process that operates on both global and local information, ensuring consistency. 
Since voxels only approximate the coarse geometry of objects and parts, overlaps frequently occur at the boundaries between adjacent parts.
The voxel initialization from the bounding boxes of the first stage may introduce noise.
To address this, we introduce a novel \textbf{\textit{voxel discarding mechanism}}, enabling precise indication of whether a voxel actually belongs to its assigned part, which aids in creating clean interfaces and allows for the efficient, \textit{simultaneous generation of all parts}. 
These fine-tuning strategies yield detailed, coherent, and generalizable part-level 3D outputs despite limited part-specific supervision.

OmniPart enables the generation of 3D objects with explicit, controllable, and semantically meaningful part structures. Extensive experiments demonstrate that our framework achieves state-of-the-art performance in part-aware 3D generation and facilitates a diverse range of downstream applications, including fine-grained compositional editing, material assignment, and animation support. Collectively, OmniPart significantly advances the creation of more interpretable and editable 3D content. Our core contributions are:
\begin{itemize}
\item A novel two-stage generative formulation and framework that strategically decouples part structure planning from part geometry synthesis, achieving superior controllability, part coherence, and overall quality.
\item An autoregressive controllable structure planning module guided by flexible 2D part masks, providing effective control over variable part granularity and 3D decomposition.
\item A spatially-conditioned part synthesis module that generates all 3D parts simultaneously and consistently, conditioned on the planned spatial layout, efficiently leveraging powerful pre-trained holistic models for high-fidelity part-aware generation with limited part-specific supervision.
\end{itemize}

\section{Related Work}
\subsection{3D Shape Generation}
DreamFusion~\cite{poole2022dreamfusion} introduces score distillation sampling to optimize 3D scenes from 2D diffusion models~\cite{rombach2022high}, enabling text-driven 3D generation without 3D supervision.
A complementary line of work~\cite{liu2023syncdreamer, long2024wonder3d, shi2023zero123plus, liu2023one2345, yang2024dreamcomposer, yang2025dreamcomposer++, xu2024instantmesh, qi2024tailor3d, zou2024triplane} generates multi-view images using 2D diffusion models and reconstructs 3D geometry via multi-view consistency algorithms. However, these 2D-driven pipelines often suffer from geometric artifacts due to inherent view inconsistency in 2D diffusion models and the lack of native 3D supervision.
To address these limitations, several methods directly model 3D data distributions using latent diffusion in geometry-aware spaces.
Approaches based on Variational Autoencoders (VAEs)~\cite{dai2017shape, wu2024direct3d, zhao2024michelangelo, zhang2024clay, li2024craftsman, lan2025ln3diff, zhang2025compress3d, hui2022neural, koo2023salad, chou2023diffusion, shim2023diffusion, li2023diffusion, li2025triposg} encode 3D shapes into latent representations, enabling efficient and effective 3D synthesis via diffusion models.
3DShape2VecSet~\cite{zhang20233dshape2vecset} introduces a cross-attention-based encoding for set-structured 3D data, while CLAY~\cite{zhang2024clay} scales latent 3D diffusion to large datasets with a hierarchical training strategy, achieving strong generation quality. TRELLIS~\cite{xiang2024structured} proposes structured latents—a unified 3D latent representation—that enables high-quality 3D object generation through a two-stage, coarse-to-fine process. And several recent methods, inspired by MeshGPT~\cite{siddiqui2024meshgpt}, condition on 3D point clouds and leverage autoregressive models to directly generate mesh faces~\cite{chen2024meshxl,chen2024meshanything,chen2024meshanythingv2,hao2024meshtron,weng2024scaling,tang2024edgerunner,liu2025freemesh,wang2025nautilus,zhao2025deepmesh}.

\subsection{Compositional Generation}
Compositional Generation has been extensively studied in the 2D domain. Several works~\cite{chatterjee2024getting,zhang2024compass,han2025spatial,huang2023t2i,huang2025t2i} focus on generating compositional images from text, enabling more controllable and structured outputs. Other methods~\cite{zhang2024transparent,pu2025art,jia2023cole} explore compositionality at the layer level, modeling images as a combination of independently generated components.
More recently, a number of works have begun to investigate 3D part generation, extending compositional reasoning into the 3D domain~\cite{chen2024comboverse,liu2024part123,chen2024partgen,li2024pasta,yan2024phycage,tang2025efficient,lin2025partcrafter}.
Comboverse~\cite{chen2024comboverse} reconstructs each object or part independently, followed by a multi-object composition stage. However, since parts are generated in isolation without global structural coherence, the final compositions often lack geometric consistency.
Part123~\cite{liu2024part123} first generates multi-view images of the object and then reconstructs the final shape using image masks.
Despite this, the reconstructed parts remain fused together, with surface-level segmentation. PASTA~\cite{li2024pasta} proposes an autoregressive transformer that represents 3D objects as sequences of cuboidal primitives and employs a transformer-based blending network to synthesize meshes.
PartGen~\cite{chen2024partgen} also adopts a multi-view pipeline: it segments and inpaints occluded regions across images before reconstructing individual parts.
However, the inconsistency among multi-view images often results in low geometric fidelity in the final 3D shapes.

\subsection{3D Part Segmentation}
Understanding the compositional structure of 3D objects through part-level segmentation is a long-standing and fundamental task in 3D vision.
Earlier research~\cite{qi2017pointnet,qi2017pointnet++,li2018pointcnn,zhao2021point,qian2022pointnext} focused primarily on designing deep architectures that learn expressive geometric representations from point clouds or meshes.
These models typically rely on fully supervised learning, which necessitates large-scale part annotations—resources that are costly to obtain and limited in coverage.
As a result, their scalability to more diverse or open-world 3D domains remains limited.
To address these limitations, recent work~\cite{liu2023partslip,yang2023sam3d,kim2024partstad,zhong2024meshsegmenter,abdelreheem2023satr,tang2024segment,thai20243x2,xue2023zerops,yang2024sampart3dsegment3dobjects,liu2024part123} has turned to 2D foundation models, including SAM~\cite{kirillov2023segment}, GLIP~\cite{li2022glip}, and CLIP~\cite{radford2021clip}, as a source of transferable visual knowledge.
These approaches typically project 3D shapes into multiple 2D views, perform segmentation in the image space, and lift the 2D masks back onto the 3D surface.
3D part segmentation can serve as a means of extracting parts from a complete shape.
While effective in segmenting visible surfaces, these pipelines inherently lack access to occluded or interior structures, leading to incomplete and view-biased segmentations.
Such limitations hinder their utility in downstream applications that demand a full understanding of object geometry. 
Furthermore, HoloPart~\cite{yang2025holopart} completes each part based on input surface masks and the overall object.
Its performance, however, is heavily dependent on the quality of the input segmentation masks.
\begin{figure*}
\centering
\includegraphics[width=\linewidth]{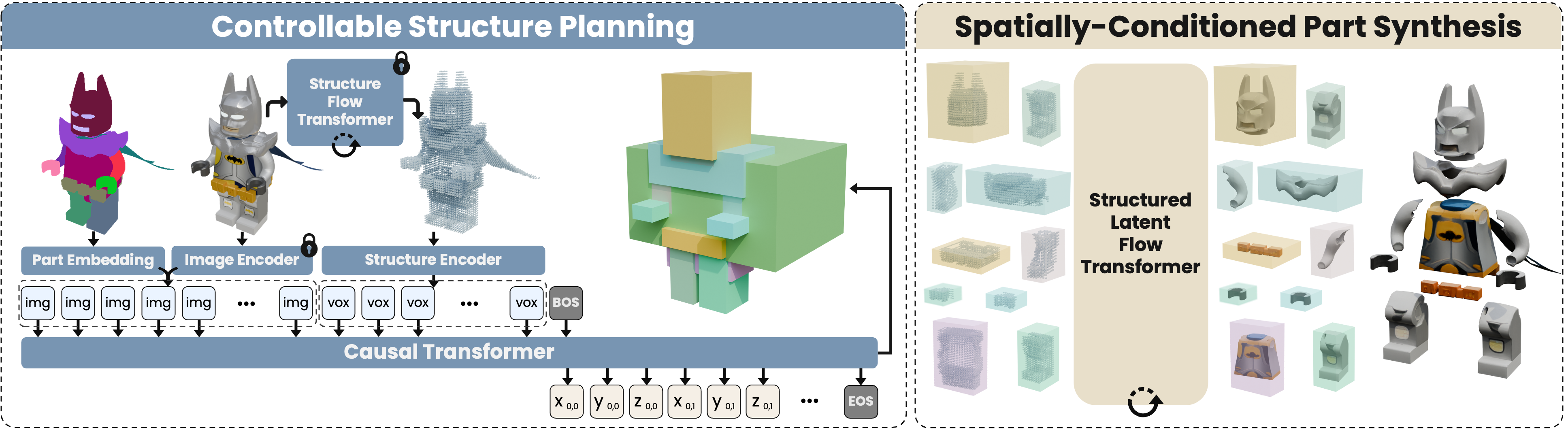}
\caption{An overview of the \textbf{OmniPart} model design. OmniPart generates part-aware, controllable, and high-quality 3D content through two key stages: part structure planning and structured part latent generation.
Built upon TRELLIS~\cite{xiang2024structured}, which provides a spatially structured sparse voxel latent space, OmniPart first predicts part-level bounding boxes via an autoregressive planner.
Then, part-specific latent codes are generated through fine-tuning of a large-scale shape model pretrained on overall objects.}
\vspace{-5pt}
\label{fig:pipeline}
\end{figure*}

\begin{figure*}
\centering
\includegraphics[width=\linewidth]{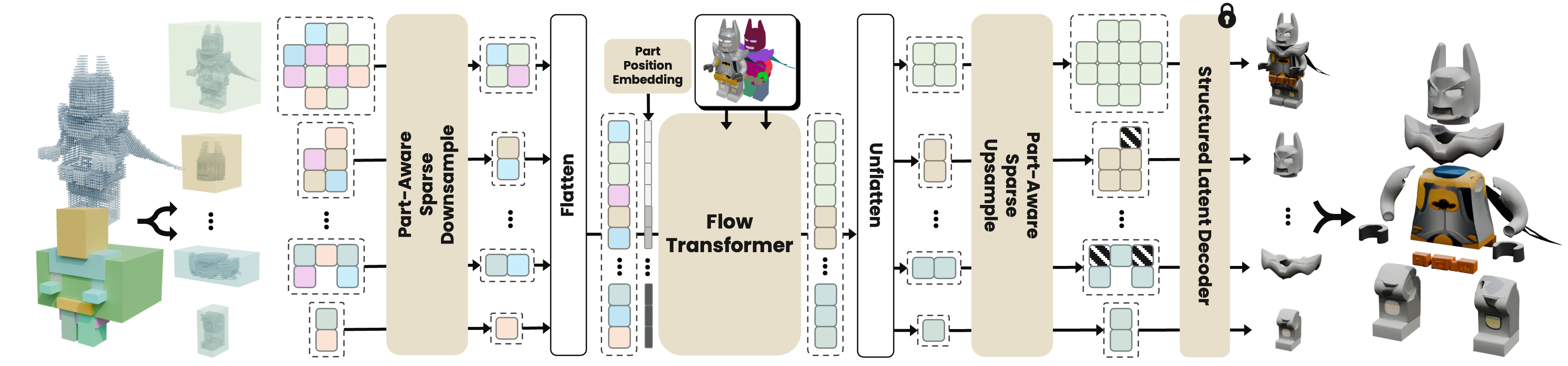}
\caption{Spatially-conditioned part synthesis. The sparse voxels of the whole shape and each part are filled with noisy latents, which are denoised with a network composed of part-aware sparse downsample/upsample layers and transformer layers. The tokens are augmented with position embeddings and part position embeddings (PPE). The denoising process also predicts a validity score for each voxel to discard redundant voxels (the ones with stripes in the figure) in each box.}
\vspace{-5pt}
\label{fig:model_second}
\end{figure*}

\section{Part-aware 3D Object Generation}
We aim to generate part-aware, controllable, and high-quality 3D content conditioned on input image with 2D part masks.
To this end, we propose OmniPart, a framework composed of two key stages: part structure planning and structured part latent generation (see \figref{fig:pipeline}).
We build on the explicit sparse voxel representation introduced in TRELLIS~\cite{xiang2024structured}, which offers a spatially structured latent space well-suited for generating coherent and semantically meaningful parts.
In Section~\ref{sec:preliminary}, we provide a brief overview of TRELLIS and its structured latent generation framework.
We then describe our Controllable Structure Planning module, which autoregressively generates part-level bounding boxes (Section~\ref{sec:part_structure_plannning}).
Next, we introduce our approach to Spatially-Conditioned Part Synthesis via fine-tuning on a large-scale pretrained model of whole-object shapes (Section~\ref{sec:structured_part_latent}). Finally, we describe the training data construction procedures for each module in Section~\ref{sec:method_data}.

\subsection{Preliminary: Generation of Structured Latent}
\label{sec:preliminary}
TRELLIS~\cite{xiang2024structured} introduces a structured latent representation. Given a 3D asset $\mathcal{C}$,  it encodes both geometric and appearance information into a unified structured latent $\mathbf{c}$, which consists of a set of local latents defined on a 3D grid:
\begin{equation}
\mathbf{c} = \left\{ (\mathbf{f}_i, \mathbf{p}_i) \right\}_{i=1}^L, \quad 
\mathbf{f}_i \in \mathbb{R}^D, \quad 
\mathbf{p}_i \in \{0, 1, \ldots, N-1\}^3,
\end{equation}
where $\mathbf{p}_i$ denotes the positional index of an active voxel in the 3D grid that intersects with the surface of $\mathcal{C}$, and $\mathbf{f}_i$ represents the local latent feature associated with that voxel. $N$ is the spatial resolution of the 3D grid. Intuitively, the active voxels $\mathbf{p}_i$ outline the coarse structure of the 3D asset, while the latents $\mathbf{f}_i$ capture finer details of its geometry and appearance. 

TRELLIS adopts a two-stage approach. In the first stage, it generates the active voxels $\mathbf{v}$ of the 3D asset $\mathcal{C}$, which cover the entire object and include its internal structure.
In the second stage, it trains a rectified flow model to generate the latent features $\mathbf{f}$ for each surface voxel.
These latents can then be decoded into meshes, NeRFs~\cite{mildenhall2021nerf}, or 3D Gaussian Splatting (3DGS)~\cite{kerbl20233d} representations.

\subsection{Controllable Structure Planning via Autoregressive Bounding Boxes Generation}
\label{sec:part_structure_plannning}
In this subsection we introduce our module design to plan the part structures of 3D objects using a coarse yet flexible representation: 3D bounding boxes. Each generated box corresponds to a meaningful part, and we employ an autoregressive generation mechanism to produce an arbitrary number of these boxes.

\noindent\textbf{Bounding box Tokenizer.} To enable autoregressive generation, we convert each 3D bounding box $\mathbf{b}$ into a discrete token sequence.  A straightforward approach is to flatten the box’s minimum and maximum coordinates into a 6n-dimensional vector, where n is the number of bounding boxes.  We then prepend a $\langle\mathrm{bos}\rangle$ (``begin-of-sequence'') token and append a $\langle\mathrm{eos}\rangle$ (``end-of-sequence'') token to clearly mark the sequence boundaries. Moreover, a predefined ordering strategy is essential to facilitate sequence modeling in bounding box generation. We sort the bounding boxes based on their minimum coordinates in z-y-x order, from lower to higher.

\noindent\textbf{Controllability and Conditions.} Since an object may consist of multiple fine-grained parts, its interpretation during planning is often ambiguous. For example, a character’s hands may be treated either as separate components or grouped together with the arms. To better control the granularity of the generated bounding boxes, we introduce a 2D mask-based control method. Given a conditional input image, 2D masks can be readily extracted using segmentation models such as SAM~\cite{kirillov2023segment}, serving as additional conditioning signals. However, since a single mask cannot encompass all object parts, there is no one-to-one correspondence between the 2D masks and the 3D bounding boxes. To address this, we adopt non-one-to-one correspondence bounding boxes, which are label-independent and eliminate the need for explicit matching between bounding boxes and 2D mask regions.

Given an input image \( I \in \mathbb{R}^{H \times W \times 3} \), we first extract visual features using DINOv2~\cite{oquab2023dinov2}:
\[
f = \mathrm{DINOv2}(I) \in \mathbb{R}^{h \times w \times d},
\]
where \( h \times w \) denotes the spatial resolution of the output feature map and \( d \) is the feature dimension.
To introduce part-aware conditioning, we resize the 2D mask \( M \in \{0, 1, \ldots, K{-}1\}^{h \times w} \), where each element \( M_{i,j} \) indicates the 2D part index at location \( (i, j) \), and \( K \) is the maximum number of parts. Correspondingly, we introduce a learnable embedding table:
\[
E \in \mathbb{R}^{K \times d}, \quad E = \texttt{nn.Embedding}(K, d),
\]
where each row \( E[k] \) provides an embedding for the \( k \)-th part label.
The final part-conditioned feature map \( f' \in \mathbb{R}^{h \times w \times d} \) is obtained by summing the visual features with the part embeddings at each spatial location:
\[
f'_{i,j} = f_{i,j} + E[M_{i,j}].
\]

To directly apply the generated bounding boxes in the Structured Part Latent Generation stage (see Sec.~\ref{sec:structured_part_latent}), we align them with the sparse voxel spatial coordinates used in that stage. 
To facilitate this alignment, we first generate voxels using the structure generation stage of TRELLIS.
The voxels are treated as point clouds and converted into fixed-length tokens $q$ using 3DShape2VecSet~\cite{zhang20233dshape2vecset} encoder. The part-conditioned feature $f'$ is flattened and concatenated with voxel tokens $q$, forming conditioning tokens as the prefix of the sequence.

\noindent\textbf{Training.} Bounding box generation can be formulated as an autoregressive sequence modeling task, making it well-suited for modern Transformer architectures. In this work, we adopt decoder-only Transformers based on the OPT~\cite{zhang2022opt} codebase as our foundation. Given the trainable parameters $\theta$ and a sequence $s$ of length $|s|$, the next-token prediction loss is defined as:
\[
\mathcal{L}_{base}(\theta) = 
- \sum_{i=1}^{|s|} \log P\left(s_{[i]} \mid s_{[1,\ldots,i-1]}; [f';\, q] \right).
\]

To ensure that each bounding box represents its corresponding object part as completely as possible, we introduce \textbf{\textit{Part Coverage Loss}}. Specifically, since our second stage model has the invalid voxel discarding ability (described below), we aim to generate relatively ``larger'' bounding boxes to ensure complete voxel coverage for each part. The coverage loss is defined as:
\begin{align}
\mathcal{L}_{\text{coverage}}(\theta) = 
&\frac{1}{|\mathcal{M}|} \Bigg(
\sum_{i \in \mathcal{M}_{\min}} 
\mathrm{ReLU}\left(s^{\text{pred}}_i - s^{\text{gt}}_i \right) \notag \\
&+ \sum_{i \in \mathcal{M}_{\max}} 
\mathrm{ReLU}\left(s^{\text{gt}}_i - s^{\text{pred}}_i \right)
\Bigg),
\end{align}
where $\mathcal{M}_{\min}$ and $\mathcal{M}_{\max}$ denote token positions corresponding to the minimum and maximum bounding box coordinates respectively. This loss penalizes bounding boxes that are too small by encouraging the predicted minimum coordinates to be smaller and the predicted maximum coordinates to be larger than the ground truth. The final loss is defined as:
\[
\mathcal{L}_{\text{total}} = \mathcal{L}_{\text{base}} + \lambda_{\text{cov}} \, \mathcal{L}_{\text{coverage}},
\]
where a scalar weight $\lambda_{\text{cov}}$ is applied to control the strength of the coverage regularization.

\subsection{Spatially-Conditioned Part Synthesis}
\label{sec:structured_part_latent}
Through part structure planning, we obtain bounding boxes that are spatially aligned with the sparse voxels. The voxels $v_i$ within the $i$-th bounding box represent the coarse geometry of the corresponding part. At this stage, our objective is to \textbf{\textit{efficiently}} generate all parts while maintaining \textbf{\textit{coherence}} across them. 
To this end, we introduce a structured part latent generation module, built upon the second stage of TRELLIS. Given the limited availability of high-quality part-level annotations, we aim to leverage the prior knowledge in the pre-trained model as much as possible.
As shown in \figref{fig:model_second}, the noisy latents of the whole shape and each part are downsampled and flattened as a 1D sequence before being fed into the transformer blocks. Here the use of whole shape tokens is to better preserve the base model prior and enhance part-to-whole coherence. 
To distinguish between different parts within the Transformer, we introduce a \textit{part position embedding} (PPE) scheme. Tokens representing the overall shape, placed at the beginning of the sequence, are assigned a shared PPE index of 0. Each subsequent part is assigned a unique PPE index starting from 1, and all tokens within the same part share the same embedding.
This design enables simultaneous denoising and decoding of all parts into consistent final representations - including meshes, NeRFs, and 3D Gaussians - in an efficient manner.

\begin{figure}
\centering
\includegraphics[width=\linewidth]{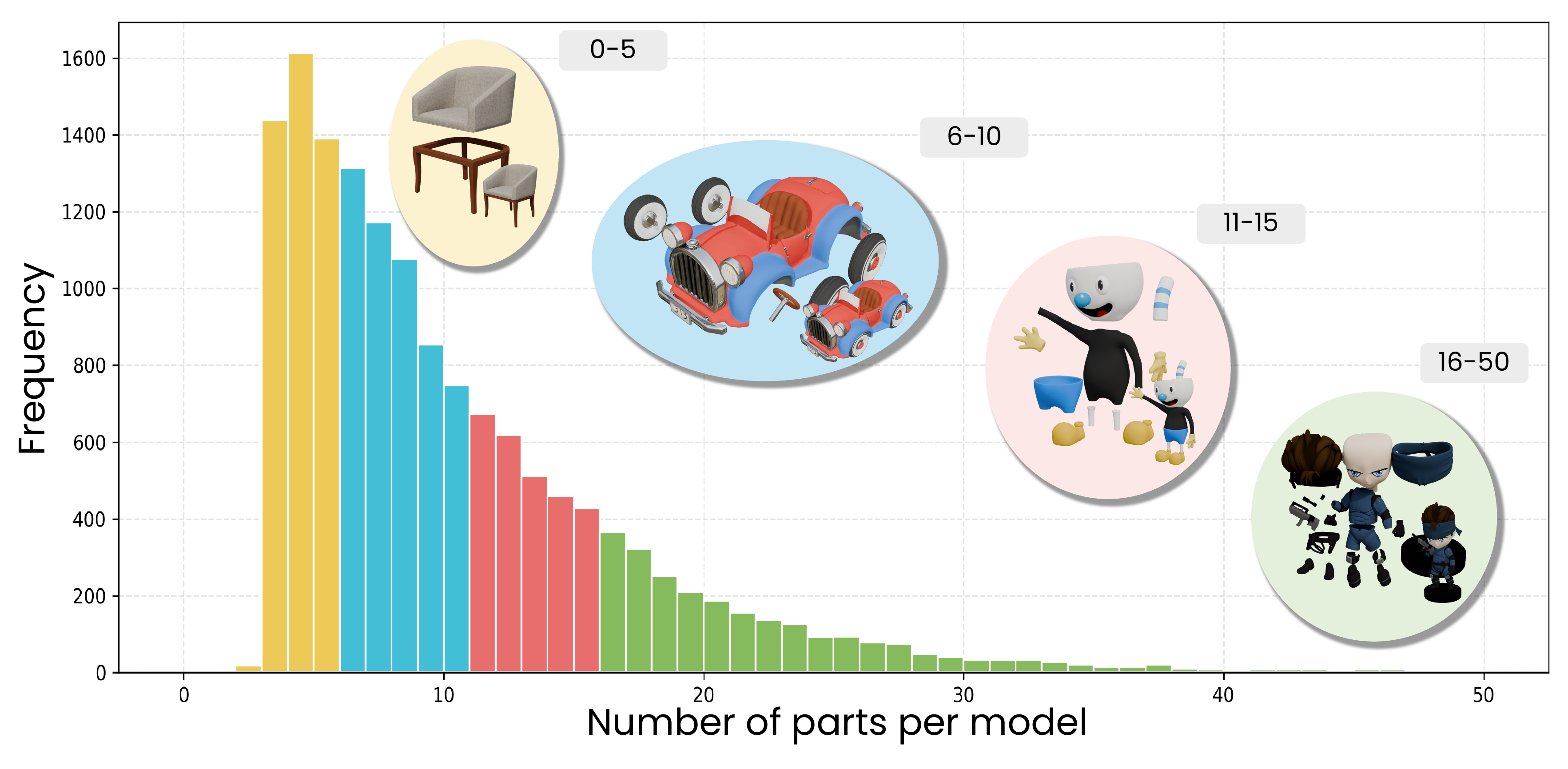}
\caption{Visualization of the training dataset. We show the distribution of part counts across the dataset (Number of parts per model vs. Frequency) and include representative examples from four different part-count ranges.}
\vspace{-10pt}
\label{fig:dataset}
\end{figure}

Since voxels can only approximate the coarse geometry of objects and parts, overlapping voxels often appear at the junctions between different parts. Moreover, directly using all voxels within a bounding box for shape initialization may introduce ``noise'' voxels that belong to other parts. To address this, we propose a \textbf{\textit{voxel discarding mechanism}} to filter out invalid voxels during initialization. We augment the original part latent with an additional dimension $\mathbf{f}_{\text{valid}}$ to indicate voxel validity. During training, we assign a value of -$\alpha$ to this new dimension for noisy voxels and +$\alpha$ for valid voxels, where $\alpha$ is a predefined constant. During inference, we determine voxel validity by applying a sigmoid activation to the additional latent dimension and comparing the result to a threshold $\beta$. A voxel is retained if its score exceeds $\beta$. In our experiments, we set $\beta$ = 0.5:
\[
\text{valid voxel} \iff \text{sigmoid}(\mathbf{f}_{\text{valid}}) > \beta.
\]

To generate structured part latents, we build on the pretrained TRELLIS model and adopt a rectified flow approach with a linear interpolation forward process:
$\mathbf{x}(t) = (1 - t)\mathbf{x}_0 + t\boldsymbol{\epsilon},$
which interpolates between the data sample $\mathbf{x}_0$ and noise $\boldsymbol{\epsilon}$ at timestep $t$.
The noise term $\boldsymbol{\epsilon}$ is approximated by a neural network $\mathbf{v}_\theta$, trained by minimizing the Conditional Flow Matching (CFM) objective~\cite{lipman2022flow}:
\[
\mathcal{L}_{\text{CFM}}(\theta) = \mathbb{E}_{t, \mathbf{x}_0, \boldsymbol{\epsilon}} \left\| \mathbf{v}_\theta(\mathbf{x}, t) - (\boldsymbol{\epsilon} - \mathbf{x}_0) \right\|_2^2.
\]

\subsection{Training Data}
\label{sec:method_data}
To train our models, we construct a dataset with part-level annotations. We begin by collecting 180K objects with part labels through a combination of filtering and manual annotation. Since the quality of part annotations varies across objects, we design a scoring system to assess labeling quality. Based on this system, 15K objects are identified as high-quality. We analyze the dataset distribution with respect to part count and present representative examples in \figref{fig:dataset}. We then construct corresponding training sets for the two modules respectively.

\noindent\textbf{Autoregressive Bounding Box Generation.}
At this stage, due to the absence of a suitable pre-trained prior, we construct training data using all 180K annotated shapes and train the autoregressive model from scratch. We extract the bounding box for each part and arrange them in a z–y–x ascending order. These boxes are then tokenized to sequences for training the autoregressive model.

\noindent\textbf{Structured Part Latent Generation.}
This stage builds on the second stage of TRELLIS for fine-tuning, using the 15K high-quality annotated shapes. Following a similar procedure to TRELLIS for constructing object-level latents, we randomly render 150 views for each part, extract features using DINOv2~\cite{oquab2023dinov2}, and unproject them onto 3D voxels for fusion. We also construct the training data that includes noise voxels. Specifically, we voxelize each part and combine them to form the full-object voxel representation. Then, we extract the voxels within each part’s bounding box as initialization, which may include noise voxels from adjacent parts.
\begin{figure*}
\centering
\includegraphics[width=0.95\linewidth]{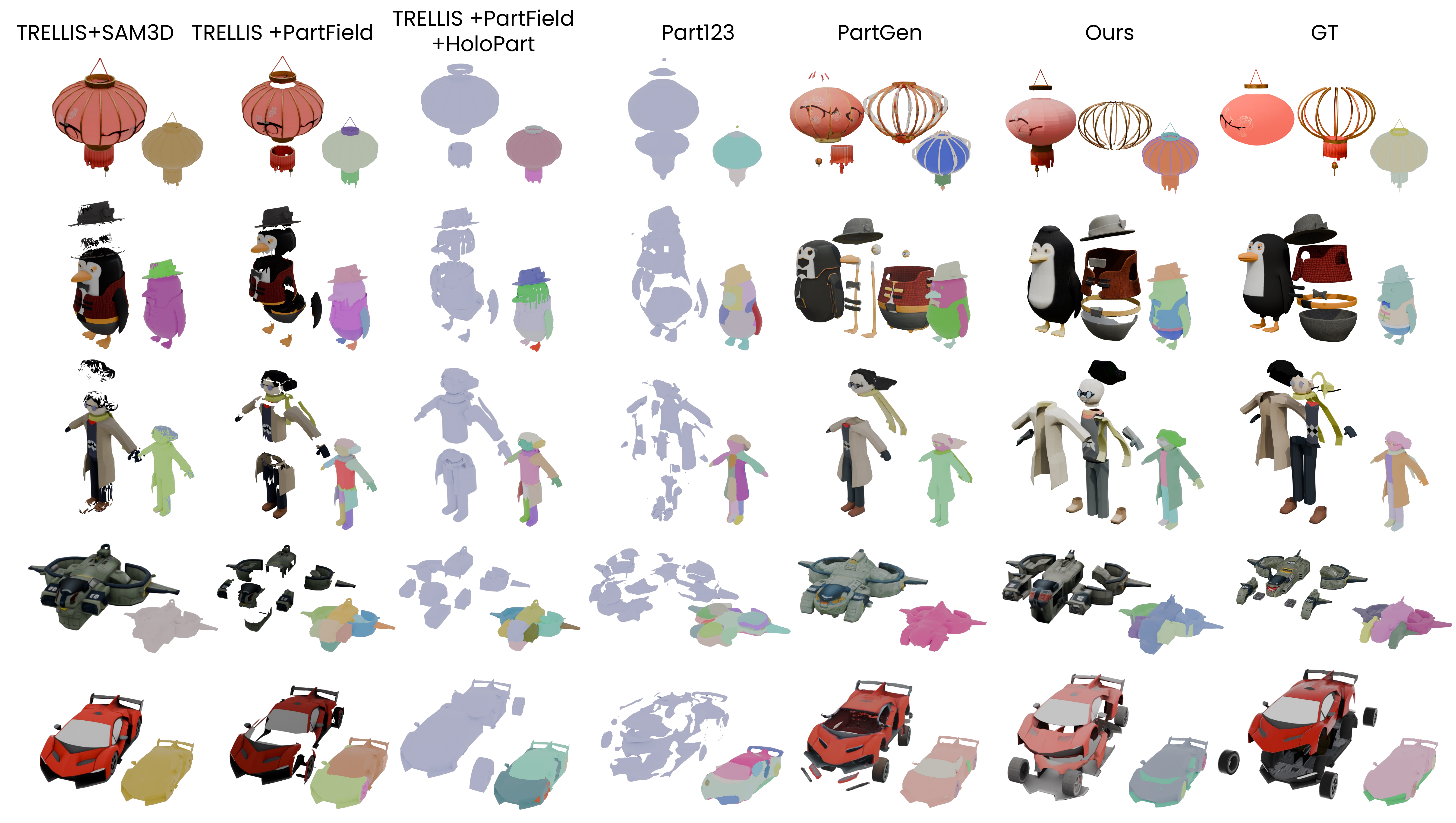}
\caption{Qualitative comparison of part-aware 3D generation.
Our method leverages TRELLIS to decode both mesh and 3D Gaussian splats, baking color onto the mesh to produce textured parts.
HoloPart and Part123 are visualized using solid colors due to the lack of texture support.
Segmentation-based methods (e.g., PartField) capture only surface-level masks, while Completion-based methods (e.g., HoloPart) are limited by segmentation quality.
PartGen generates full parts but with low geometric and semantic quality.
In contrast, our method achieves low semantic coupling and high structural cohesion.}
\label{fig:exp_part}
\end{figure*}

\begin{figure*}
\centering
\includegraphics[width=0.92\linewidth]{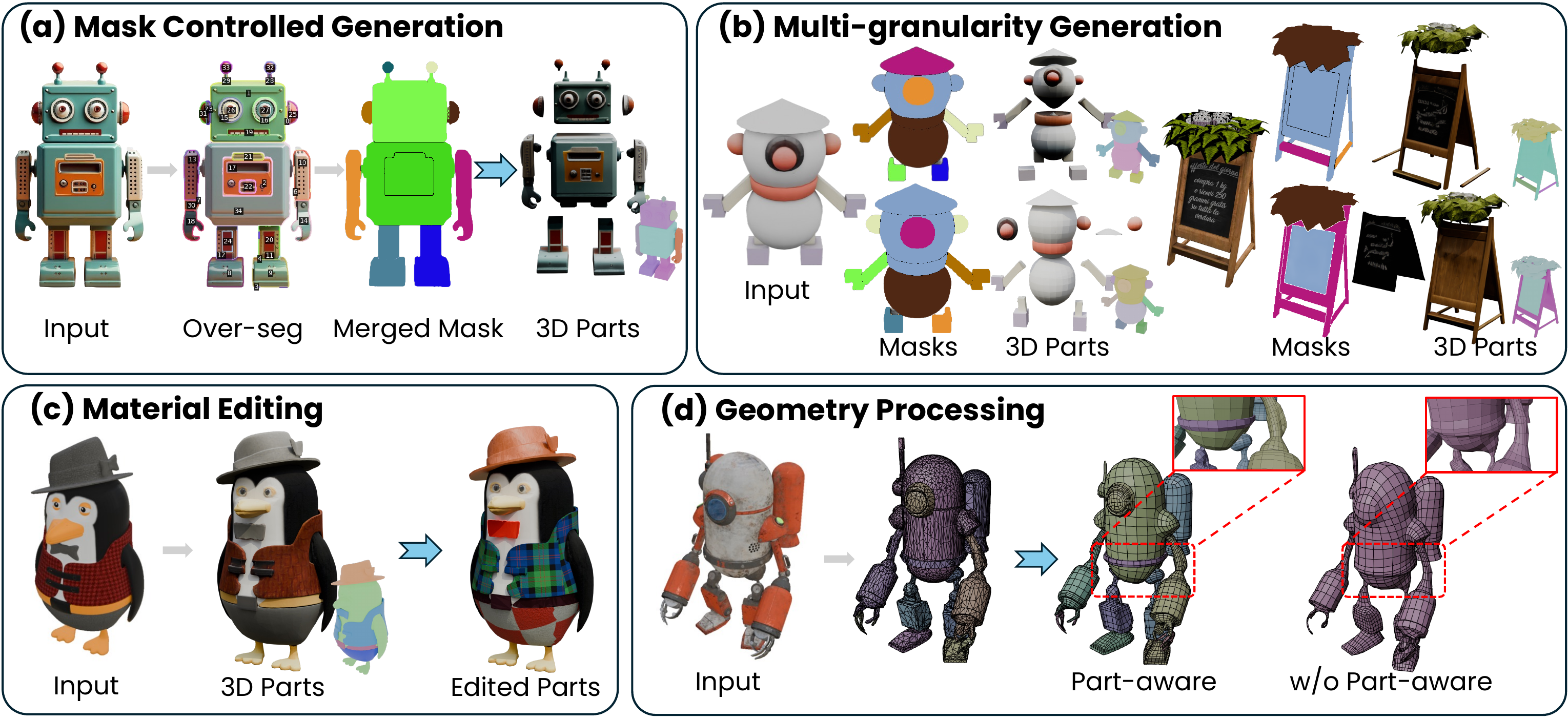}
\caption{Applications of our part-aware 3D generation framework.
(a) \textbf{Mask-Controlled Generation}: Users can specify 2D masks to guide the structure of the generated parts.
(b) \textbf{Multi-Granularity Generation}: Adjusting the segmentation scale of 2D masks enables generation at different levels of part granularity.
(c) \textbf{Material Editing}: Part-specific textures, such as clothing items, can be modified independently.
(d) \textbf{Geometry Processing}: Our part-aware outputs support high-quality geometry processing (such as remeshing) and preserve structural coherence, avoiding artifacts at part boundaries.}
\label{fig:application}
\end{figure*}

\section{Experiments}
\noindent\textbf{Evaluation Protocol.} To evaluate our model, we construct a test set of 300 objects sampled from the dataset described in \secref{sec:method_data}.
Based on the number of parts, we divide the objects into four groups: 0–5, 6–10, 11–15, and 16–50 parts.
Objects were selected proportionally from each group to ensure both part count diversity and category coverage.
We first evaluate the bounding box generation (planning) stage of our model in \secref{sec:exp_bbox}.
Next, we assess the full pipeline for part-aware 3D content generation in \secref{sec:exp_part_generation}.
Finally, we showcase practical applications of our method in \secref{sec:exp_applications}.
\subsection{Bounding Box Generation}
\label{sec:exp_bbox}
\noindent\textbf{Metrics and Baselines.} To evaluate our model’s performance in the bounding box generation stage, we introduce three metrics: \textit{BBox IoU}, \textit{Voxel recall}, and \textit{Voxel IoU}.
BBox IoU directly measures the overlap between the predicted bounding box and the ground truth.
Since the primary goal of this stage is to provide coarse part voxels for the subsequent latent generation stage, we also introduce two voxel-level metrics.
Voxel recall assesses the proportion of valid part voxels that fall within the predicted bounding box—important because the Spatially-Conditioned Part Synthesis stage can discard extraneous voxels but cannot recover missing ones.
Voxel IoU computes the intersection-over-union between the voxels inside the predicted and ground-truth bounding boxes, providing a global measure of voxel-level overlap. 
Since our predicted bounding boxes are not constrained by semantic labels or quantity, there is no one-to-one correspondence between ground truth and predictions.
To compute the metrics, we match each ground-truth bounding box with its proximate predicted counterpart.

At this stage, we obtain the object’s overall coarse voxels from input images and aim to plan the bounding boxes for individual parts.
PartField~\cite{liu2025partfield}, which supports zero-shot point cloud segmentation, serves as a baseline for comparison.
We treat the voxel representation as a point cloud and feed it into PartField to obtain per-point segmentation features.
As PartField requires the number of parts to define the segmentation scale, we provide the ground-truth number of parts as input.
After segmenting the point cloud, we extract the bounding box of each predicted part and use these to compute evaluation metrics.

\begin{table}
\centering

\resizebox{0.95\linewidth}{!}{

\begin{tabular}{c|ccc}
\hline
 Method  & Voxel recall $\uparrow$ & Voxel IoU $\uparrow$ & Bbox IoU $\uparrow$ \\  
 \hline  
 PartField~\cite{liu2025partfield} & 79.12 & 39.02 & 27.30  \\
 OmniPart (w/o 2D mask) & 66.98 & 31.44 & 25.90 \\
 OmniPart (w/o coverage loss) & 64.50 & 50.56 & \textbf{41.24} \\
 OmniPart (Ours) & \textbf{85.96} & \textbf{61.02} & 38.37  \\ 
    \hline
\end{tabular}
}
\caption{Quantitative results for bounding box generation (\%). We report BBox IoU to measure bounding box overlap, Voxel Recall to assess coverage of valid part voxels, and Voxel IoU to quantify overall voxel-level consistency between predictions and ground truth.}
\vspace{-15pt}
\label{tab:exp_bbox}
\end{table}

\noindent\textbf{Results and Ablation Analysis.} As shown in the first and last rows of \tabref{tab:exp_bbox}, our results significantly exceed the results of PartField. It means that the accuracy and integrity of our predicted bounding box are better than the baseline model. 

We also conduct ablation experiments to assess the influence of the \textit{\textbf{coverage loss}} and \textit{\textbf{2D mask}} on bounding box generation.
Specifically, we train one model without coverage loss and another without the 2D mask input, and evaluate their performance.
As shown in \tabref{tab:exp_bbox}, although the model without coverage loss produces bounding boxes that appear closer to the ground truth, its voxel-level recall and IoU are significantly lower, which negatively impacts the performance of the second stage.
And the model without the 2D mask input fails to effectively control the size and placement of the generated bounding boxes.

\begin{table}
\centering

\resizebox{0.95\linewidth}{!}{

\centering
\begin{tabular}{c|ccc|ccc}
\hline
\multirow{2}{*}{Method} & \multicolumn{3}{c|}{Part-level} & \multicolumn{3}{c}{Overall-object} \\
                        & CD $\downarrow$ & F1-0.1 $\uparrow$ & F1-0.5 $\uparrow$ & CD $\downarrow$ & F1-0.1 $\uparrow$ & F1-0.5 $\uparrow$ \\
 \hline  
 TRELLIS+SAM3D & 0.49 & 0.38 & 0.28 & 0.08 & 0.92 & 0.77 \\
 TRELLIS+PartField & 0.19 & 0.69 & 0.52 & 0.08 & 0.92 & 0.77 \\
 TRELLIS+PartField+HoloPart & 0.19 & 0.68 & 0.51 & 0.08 & 0.91 & 0.77 \\
 Part123 & 0.43 & 0.31 & 0.16 & 0.47 & 0.33 &  0.19 \\
 PartGen & 0.44  & 0.43 & 0.30 & 0.11  &  0.86 & 0.69 \\
 OmniPart (Ours) & \textbf{0.18} & \textbf{0.74} & \textbf{0.59}  & \textbf{0.07} & \textbf{0.93} & \textbf{0.80} \\ 
    \hline
\end{tabular}
}
\caption{Quantitative results of part-level and whole-object generation (\%).
We report Chamfer Distance (CD) and F1-score after normalizing all shapes to a unified scale within [-0.5, 0.5].
F1-score is computed at two thresholds (CD $<$ 0.1 and CD $<$ 0.05) to capture both coarse and fine geometric accuracy.
To account for orientation differences, each object is evaluated under four rotations (0°, 90°, 180°, 270°), and the best score is reported.}
\vspace{-15pt}
\label{tab:exp_part}
\end{table}
\begin{table}
\centering

\resizebox{0.95\linewidth}{!}{

\begin{tabular}{c|ccc}
\hline
 Method  & Part123~\cite{liu2024part123} & PartGen~\cite{chen2024partgen} & Ours \\  
 Time (minute) & $\sim$
 15 & $\sim$ 5 & $\sim$ 0.75 \\
    \hline
\end{tabular}
}
\caption{End-to-end generation time comparison from a single image to part-level 3D outputs.}
\vspace{-18pt}
\label{tab:exp_time}
\end{table}

\subsection{Part-Aware 3D Content Generation}
\label{sec:exp_part_generation}
\noindent\textbf{Metrics and Baselines.}
We consider three primary approaches for part-aware 3D content generation.
The first approach generates a complete 3D shape and then segments it into distinct parts.
The second builds upon this by completing each part individually after segmentation.
The third directly infers part-aware 3D shapes from a single image.
For the first approach, we adopt TRELLIS+SAM3D~\cite{yang2023sam3d} and TRELLIS+PartField~\cite{liu2025partfield} as baseline methods.
For the second, we use TRELLIS+Partfield+HoloPart~\cite{yang2025holopart}.
For the third, we compare against Part123~\cite{liu2024part123} and PartGen~\cite{chen2024partgen}.

To evaluate both the quality of individual generated parts and the overall coherence of the merged object, we report metrics at two levels: part-level and whole-object.
Specifically, we normalize both the ground-truth and predicted overall shapes to a unified scale within [-0.5, 0.5] before computing Chamfer Distance (CD) and F1-score for each level.
For the F1-score, we use two distance thresholds—CD $<$ 0.1 and CD $<$ 0.05—to assess coarse- and fine-level geometric alignment. Due to the varying orientations of objects generated by different methods, we rotate each object by 0, 90, 180, and 270 degrees and evaluate the results under each rotation.
We report the highest score obtained across these orientations.

\noindent\textbf{Comparison Results.} We present quantitative results in \tabref{tab:exp_part}.
As shown in the last three rows of \tabref{tab:exp_part}, our method significantly outperforms the part-aware generation baselines at both the part-level and whole-object level.
It is worth noting that the first two rows under the ``overall-object'' category correspond to results directly generated by TRELLIS without part decomposition.
In contrast, our merged full-object shapes achieve higher performance, as our method can generate complete geometry for each part, including the boundaries and occluded regions—areas that TRELLIS alone cannot accurately reconstruct when generating the object as a whole.

The qualitative results are shown in \figref{fig:exp_part}. Our method leverages TRELLIS’s ability to decode both mesh and 3D Gaussian splats simultaneously.
We bake the color information from the 3D Gaussians onto the mesh surface, thereby generating textured parts.
For Holopart and Part123, we use the official implementations.
Since their generated meshes do not contain textures, we visualize them using solid colors. 
Segmentation-based methods can only produce surface-level masks and fail to recover complete part geometry.
HoloPart relies on a completion strategy but remains constrained by the quality of the initial segmentation.
While PartGen is capable of generating complete parts, its outputs often suffer from low geometric fidelity and limited semantic plausibility.
In contrast, our method produces results with low semantic coupling and high structural cohesion, ensuring both part-level independence and global consistency.

\begin{figure*}
\centering
\includegraphics[width=\linewidth]{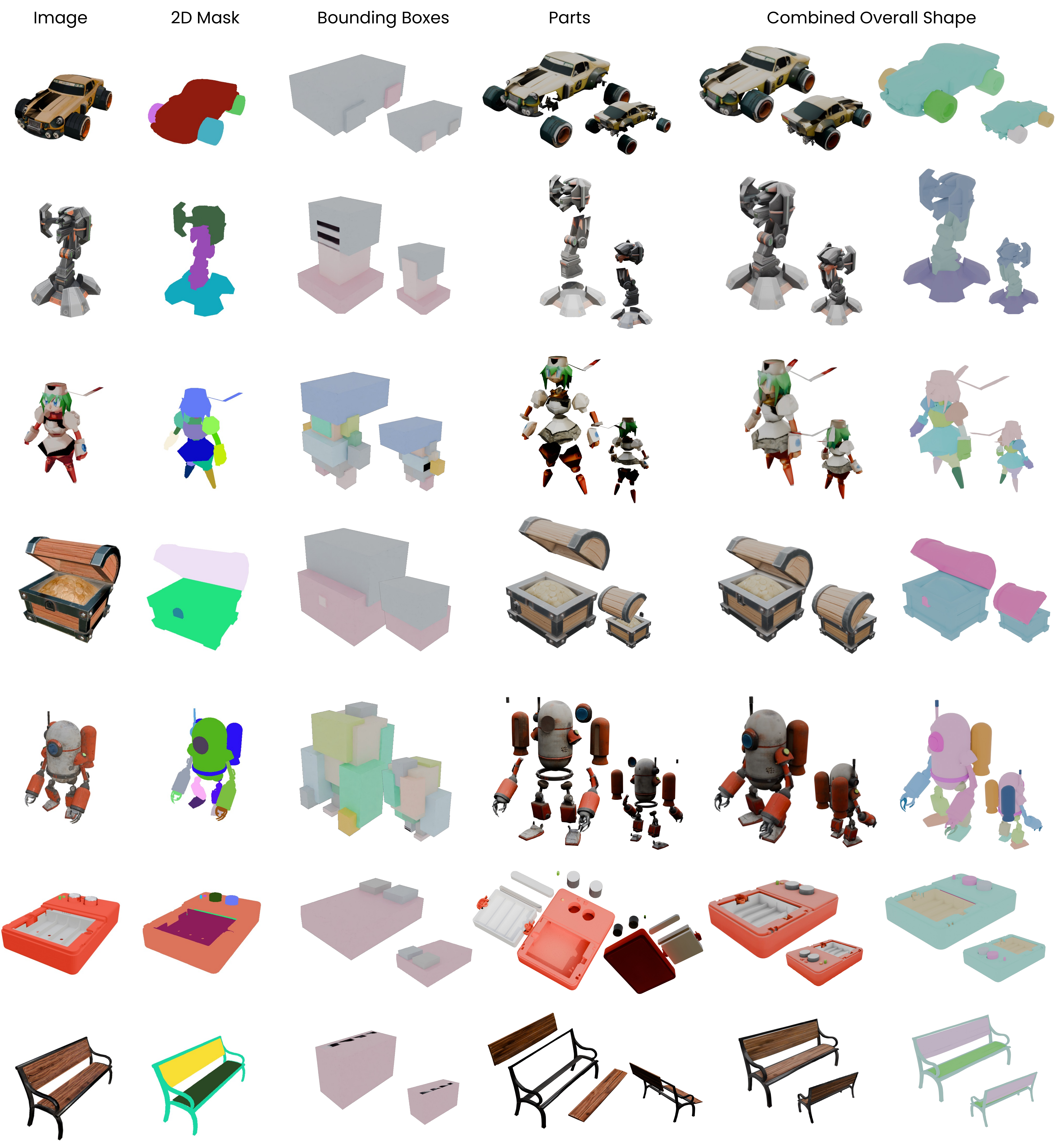}
\caption{Qualitative results of our complete pipeline. We show the input image and 2D masks, along with the generated bounding boxes, individually generated part meshes, and the combined full-object mesh. As illustrated, our method enables precise control over part granularity via 2D masks and produces high-quality geometry and texture. The generated 3D parts exhibit low semantic entanglement and high structural cohesion, demonstrating the effectiveness of our part-aware 3D content generation.}
\label{fig:our_results}
\end{figure*}

\noindent\textbf{Pipeline Results.}
We present the results of our complete pipeline in \figref{fig:our_results}.
The figure shows the input image and 2D masks, along with the generated bounding boxes, individually generated part meshes, and the combined full-object mesh.
As illustrated, our method enables precise control over part granularity via 2D masks and generates high-quality geometry and texture.

\noindent\textbf{Efficiency.}
We aim to make part-aware 3D content generation efficient.
Part123 requires generating multiple views, optimizing the reconstruction process, and then obtaining the segmentation results.
PartGen first generates multi-view images and performs segmentation on them to obtain multi-view part masks.
These segmented views are then completed and used to reconstruct each part in 3D.
In contrast, our unified backbone design enables simultaneous generation of all parts and supports direct decoding into mesh, 3D Gaussian splats (3DGS), or NeRF representations.
We compare the end-to-end generation time—from a single image to part-level 3D outputs—against both methods, and as shown in Table~\ref{tab:exp_time}, our approach demonstrates substantial efficiency gains. 

\subsection{Applications}
\label{sec:exp_applications}
Our method generates high-quality part-aware 3D content directly from a single input image and naturally supports a range of downstream applications, including \textit{\textbf{Animation}}, \textit{\textbf{Mask-Controlled Generation}}, \textit{\textbf{Multi-Granularity Generation}}, \textit{\textbf{Material Editing}}, and \textit{\textbf{Geometry Processing}}, as illustrated in \figref{fig:application}.
We showcase the animation of our generated meshes in the video.

\noindent\textbf{Mask-Controlled Generation.} 
We can easily control the generation of 3D parts by specifying a 2D mask.
To facilitate this, we design a simple and efficient process to obtain accurate 2D masks by merging over-segmented regions produced by SAM.
An example is shown in \figref{fig:application} (a), where a 2D mask is used to control the structure of the generated robots.

\noindent\textbf{Multi-Granularity Generation.}
We achieve multi-granularity 3D part generation by controlling the granularity of the 2D segmentation masks.
As shown in \figref{fig:application}(b), we generate 3D objects with different part granularities by adjusting the segmentation scale of the 2D masks for the robot and the sketchpad.

\noindent\textbf{Material Editing.}
With the generated part-aware 3D objects, material properties can be easily edited at the part level.
For example, in \figref{fig:application}(c), we modify the textures of the penguin’s hat, tie, clothes, and pants.
This enables designers to freely customize object appearance with fine-grained control.

\noindent\textbf{Geometry Processing.}
With the part-aware generated 3D objects, geometry processing becomes significantly more convenient and effective.
For example, in \figref{fig:application}(d), we convert the generated mesh from triangles to quadrilaterals using a remeshing tool.
Our part-aware representations handle each part cleanly and consistently, whereas non-part-aware results often exhibit artifacts at the junctions between parts.

\section{Conclusion and Discussion}
\noindent\textbf{Conclusion.} We present OmniPart, a part-aware 3D generation framework that addresses the limitations of traditional monolithic 3D models by explicitly modeling semantically meaningful and structurally coherent object parts.
Our two-stage pipeline first plans a controllable part-level spatial layout via autoregressive bounding box prediction, guided by flexible 2D masks.
Then, conditioned on this layout, a spatially-aware synthesis module generates high-quality 3D parts simultaneously and efficiently, leveraging a pretrained holistic generator.
Despite limited part-level supervision, OmniPart produces detailed and coherent part structures with low semantic coupling and high structural cohesion.
It enables precise control over part granularity and supports a wide range of applications, including animation, material editing, and geometric processing.
Extensive experiments confirm its state-of-the-art performance and practical versatility, advancing the creation of interpretable, editable 3D assets for modern visual computing.

\noindent\textbf{Limitation.} To simplify training in the first stage, we currently use axis-aligned bounding boxes.
However, in some cases, this may result in excessive inclusion of noisy voxels being passed to the second stage.
Exploring more precise representations for structure planning remains a promising direction for future work.

\bibliographystyle{ACM-Reference-Format}
\bibliography{sample-bibliography}


\begin{thebibliography}{76}


\ifx \showCODEN    \undefined \def \showCODEN     #1{\unskip}     \fi
\ifx \showDOI      \undefined \def \showDOI       #1{#1}\fi
\ifx \showISBNx    \undefined \def \showISBNx     #1{\unskip}     \fi
\ifx \showISBNxiii \undefined \def \showISBNxiii  #1{\unskip}     \fi
\ifx \showISSN     \undefined \def \showISSN      #1{\unskip}     \fi
\ifx \showLCCN     \undefined \def \showLCCN      #1{\unskip}     \fi
\ifx \shownote     \undefined \def \shownote      #1{#1}          \fi
\ifx \showarticletitle \undefined \def \showarticletitle #1{#1}   \fi
\ifx \showURL      \undefined \def \showURL       {\relax}        \fi
\providecommand\bibfield[2]{#2}
\providecommand\bibinfo[2]{#2}
\providecommand\natexlab[1]{#1}
\providecommand\showeprint[2][]{arXiv:#2}

\bibitem[Abdelreheem et~al\mbox{.}(2023)]%
        {abdelreheem2023satr}
\bibfield{author}{\bibinfo{person}{Ahmed Abdelreheem}, \bibinfo{person}{Ivan Skorokhodov}, \bibinfo{person}{Maks Ovsjanikov}, {and} \bibinfo{person}{Peter Wonka}.} \bibinfo{year}{2023}\natexlab{}.
\newblock \showarticletitle{Satr: Zero-shot semantic segmentation of 3d shapes}. In \bibinfo{booktitle}{\emph{ICCV}}.
\newblock


\bibitem[Chatterjee et~al\mbox{.}(2024)]%
        {chatterjee2024getting}
\bibfield{author}{\bibinfo{person}{Agneet Chatterjee}, \bibinfo{person}{Gabriela Ben~Melech Stan}, \bibinfo{person}{Estelle Aflalo}, \bibinfo{person}{Sayak Paul}, \bibinfo{person}{Dhruba Ghosh}, \bibinfo{person}{Tejas Gokhale}, \bibinfo{person}{Ludwig Schmidt}, \bibinfo{person}{Hannaneh Hajishirzi}, \bibinfo{person}{Vasudev Lal}, \bibinfo{person}{Chitta Baral}, {et~al\mbox{.}}} \bibinfo{year}{2024}\natexlab{}.
\newblock \showarticletitle{Getting it right: Improving spatial consistency in text-to-image models}. In \bibinfo{booktitle}{\emph{ECCV}}.
\newblock


\bibitem[Chen et~al\mbox{.}(2024b)]%
        {chen2024partgen}
\bibfield{author}{\bibinfo{person}{Minghao Chen}, \bibinfo{person}{Roman Shapovalov}, \bibinfo{person}{Iro Laina}, \bibinfo{person}{Tom Monnier}, \bibinfo{person}{Jianyuan Wang}, \bibinfo{person}{David Novotny}, {and} \bibinfo{person}{Andrea Vedaldi}.} \bibinfo{year}{2024}\natexlab{b}.
\newblock \showarticletitle{PartGen: Part-level 3D Generation and Reconstruction with Multi-View Diffusion Models}.
\newblock \bibinfo{journal}{\emph{arXiv preprint arXiv:2412.18608}} (\bibinfo{year}{2024}).
\newblock


\bibitem[Chen et~al\mbox{.}(2024a)]%
        {chen2024meshxl}
\bibfield{author}{\bibinfo{person}{Sijin Chen}, \bibinfo{person}{Xin Chen}, \bibinfo{person}{Anqi Pang}, \bibinfo{person}{Xianfang Zeng}, \bibinfo{person}{Wei Cheng}, \bibinfo{person}{Yijun Fu}, \bibinfo{person}{Fukun Yin}, \bibinfo{person}{Billzb Wang}, \bibinfo{person}{Jingyi Yu}, \bibinfo{person}{Gang Yu}, {et~al\mbox{.}}} \bibinfo{year}{2024}\natexlab{a}.
\newblock \showarticletitle{Meshxl: Neural coordinate field for generative 3d foundation models}. In \bibinfo{booktitle}{\emph{NeurIPS}}.
\newblock


\bibitem[Chen et~al\mbox{.}(2025)]%
        {chen2024meshanything}
\bibfield{author}{\bibinfo{person}{Yiwen Chen}, \bibinfo{person}{Tong He}, \bibinfo{person}{Di Huang}, \bibinfo{person}{Weicai Ye}, \bibinfo{person}{Sijin Chen}, \bibinfo{person}{Jiaxiang Tang}, \bibinfo{person}{Xin Chen}, \bibinfo{person}{Zhongang Cai}, \bibinfo{person}{Lei Yang}, \bibinfo{person}{Gang Yu}, {et~al\mbox{.}}} \bibinfo{year}{2025}\natexlab{}.
\newblock \showarticletitle{Meshanything: Artist-created mesh generation with autoregressive transformers}. In \bibinfo{booktitle}{\emph{ICLR}}.
\newblock


\bibitem[Chen et~al\mbox{.}(2024d)]%
        {chen2024comboverse}
\bibfield{author}{\bibinfo{person}{Yongwei Chen}, \bibinfo{person}{Tengfei Wang}, \bibinfo{person}{Tong Wu}, \bibinfo{person}{Xingang Pan}, \bibinfo{person}{Kui Jia}, {and} \bibinfo{person}{Ziwei Liu}.} \bibinfo{year}{2024}\natexlab{d}.
\newblock \showarticletitle{Comboverse: Compositional 3d assets creation using spatially-aware diffusion guidance}. In \bibinfo{booktitle}{\emph{ECCV}}.
\newblock


\bibitem[Chen et~al\mbox{.}(2024c)]%
        {chen2024meshanythingv2}
\bibfield{author}{\bibinfo{person}{Yiwen Chen}, \bibinfo{person}{Yikai Wang}, \bibinfo{person}{Yihao Luo}, \bibinfo{person}{Zhengyi Wang}, \bibinfo{person}{Zilong Chen}, \bibinfo{person}{Jun Zhu}, \bibinfo{person}{Chi Zhang}, {and} \bibinfo{person}{Guosheng Lin}.} \bibinfo{year}{2024}\natexlab{c}.
\newblock \showarticletitle{Meshanything v2: Artist-created mesh generation with adjacent mesh tokenization}.
\newblock \bibinfo{journal}{\emph{arXiv preprint arXiv:2408.02555}} (\bibinfo{year}{2024}).
\newblock


\bibitem[Chou et~al\mbox{.}(2023)]%
        {chou2023diffusion}
\bibfield{author}{\bibinfo{person}{Gene Chou}, \bibinfo{person}{Yuval Bahat}, {and} \bibinfo{person}{Felix Heide}.} \bibinfo{year}{2023}\natexlab{}.
\newblock \showarticletitle{Diffusion-sdf: Conditional generative modeling of signed distance functions}. In \bibinfo{booktitle}{\emph{ICCV}}.
\newblock


\bibitem[Dai et~al\mbox{.}(2017)]%
        {dai2017shape}
\bibfield{author}{\bibinfo{person}{Angela Dai}, \bibinfo{person}{Charles Ruizhongtai~Qi}, {and} \bibinfo{person}{Matthias Nie{\ss}ner}.} \bibinfo{year}{2017}\natexlab{}.
\newblock \showarticletitle{Shape completion using 3d-encoder-predictor cnns and shape synthesis}. In \bibinfo{booktitle}{\emph{CVPR}}.
\newblock


\bibitem[Han et~al\mbox{.}(2025)]%
        {han2025spatial}
\bibfield{author}{\bibinfo{person}{Woojung Han}, \bibinfo{person}{Yeonkyung Lee}, \bibinfo{person}{Chanyoung Kim}, \bibinfo{person}{Kwanghyun Park}, {and} \bibinfo{person}{Seong~Jae Hwang}.} \bibinfo{year}{2025}\natexlab{}.
\newblock \showarticletitle{Spatial Transport Optimization by Repositioning Attention Map for Training-Free Text-to-Image Synthesis}.
\newblock \bibinfo{journal}{\emph{arXiv preprint arXiv:2503.22168}} (\bibinfo{year}{2025}).
\newblock


\bibitem[Hao et~al\mbox{.}(2024)]%
        {hao2024meshtron}
\bibfield{author}{\bibinfo{person}{Zekun Hao}, \bibinfo{person}{David~W Romero}, \bibinfo{person}{Tsung-Yi Lin}, {and} \bibinfo{person}{Ming-Yu Liu}.} \bibinfo{year}{2024}\natexlab{}.
\newblock \showarticletitle{Meshtron: High-Fidelity, Artist-Like 3D Mesh Generation at Scale}.
\newblock \bibinfo{journal}{\emph{arXiv preprint arXiv:2412.09548}} (\bibinfo{year}{2024}).
\newblock


\bibitem[Huang et~al\mbox{.}(2025)]%
        {huang2025t2i}
\bibfield{author}{\bibinfo{person}{Kaiyi Huang}, \bibinfo{person}{Chengqi Duan}, \bibinfo{person}{Kaiyue Sun}, \bibinfo{person}{Enze Xie}, \bibinfo{person}{Zhenguo Li}, {and} \bibinfo{person}{Xihui Liu}.} \bibinfo{year}{2025}\natexlab{}.
\newblock \showarticletitle{T2I-CompBench++: An Enhanced and Comprehensive Benchmark for Compositional Text-to-Image Generation}.
\newblock \bibinfo{journal}{\emph{IEEE Transactions on Pattern Analysis and Machine Intelligence}} (\bibinfo{year}{2025}).
\newblock


\bibitem[Huang et~al\mbox{.}(2023)]%
        {huang2023t2i}
\bibfield{author}{\bibinfo{person}{Kaiyi Huang}, \bibinfo{person}{Kaiyue Sun}, \bibinfo{person}{Enze Xie}, \bibinfo{person}{Zhenguo Li}, {and} \bibinfo{person}{Xihui Liu}.} \bibinfo{year}{2023}\natexlab{}.
\newblock \showarticletitle{T2i-compbench: A comprehensive benchmark for open-world compositional text-to-image generation}. In \bibinfo{booktitle}{\emph{NeurIPS}}.
\newblock


\bibitem[Hui et~al\mbox{.}(2022)]%
        {hui2022neural}
\bibfield{author}{\bibinfo{person}{Ka-Hei Hui}, \bibinfo{person}{Ruihui Li}, \bibinfo{person}{Jingyu Hu}, {and} \bibinfo{person}{Chi-Wing Fu}.} \bibinfo{year}{2022}\natexlab{}.
\newblock \showarticletitle{Neural wavelet-domain diffusion for 3d shape generation}. In \bibinfo{booktitle}{\emph{SIGGRAPH Asia}}.
\newblock


\bibitem[Jia et~al\mbox{.}(2023)]%
        {jia2023cole}
\bibfield{author}{\bibinfo{person}{Peidong Jia}, \bibinfo{person}{Chenxuan Li}, \bibinfo{person}{Yuhui Yuan}, \bibinfo{person}{Zeyu Liu}, \bibinfo{person}{Yichao Shen}, \bibinfo{person}{Bohan Chen}, \bibinfo{person}{Xingru Chen}, \bibinfo{person}{Yinglin Zheng}, \bibinfo{person}{Dong Chen}, \bibinfo{person}{Ji Li}, {et~al\mbox{.}}} \bibinfo{year}{2023}\natexlab{}.
\newblock \showarticletitle{COLE: A Hierarchical Generation Framework for Multi-Layered and Editable Graphic Design}.
\newblock \bibinfo{journal}{\emph{arXiv preprint arXiv:2311.16974}} (\bibinfo{year}{2023}).
\newblock


\bibitem[Kerbl et~al\mbox{.}(2023)]%
        {kerbl20233d}
\bibfield{author}{\bibinfo{person}{Bernhard Kerbl}, \bibinfo{person}{Georgios Kopanas}, \bibinfo{person}{Thomas Leimk{\"u}hler}, {and} \bibinfo{person}{George Drettakis}.} \bibinfo{year}{2023}\natexlab{}.
\newblock \showarticletitle{3d gaussian splatting for real-time radiance field rendering.}
\newblock \bibinfo{journal}{\emph{ACM Trans. Graph.}} \bibinfo{volume}{42}, \bibinfo{number}{4} (\bibinfo{year}{2023}), \bibinfo{pages}{139--1}.
\newblock


\bibitem[Kim and Sung(2024)]%
        {kim2024partstad}
\bibfield{author}{\bibinfo{person}{Hyunjin Kim} {and} \bibinfo{person}{Minhyuk Sung}.} \bibinfo{year}{2024}\natexlab{}.
\newblock \showarticletitle{PartSTAD: 2D-to-3D Part Segmentation Task Adaptation}. In \bibinfo{booktitle}{\emph{ECCV}}.
\newblock


\bibitem[Kirillov et~al\mbox{.}(2023)]%
        {kirillov2023segment}
\bibfield{author}{\bibinfo{person}{Alexander Kirillov}, \bibinfo{person}{Eric Mintun}, \bibinfo{person}{Nikhila Ravi}, \bibinfo{person}{Hanzi Mao}, \bibinfo{person}{Chloe Rolland}, \bibinfo{person}{Laura Gustafson}, \bibinfo{person}{Tete Xiao}, \bibinfo{person}{Spencer Whitehead}, \bibinfo{person}{Alexander~C Berg}, \bibinfo{person}{Wan-Yen Lo}, {et~al\mbox{.}}} \bibinfo{year}{2023}\natexlab{}.
\newblock \showarticletitle{Segment anything}. In \bibinfo{booktitle}{\emph{ICCV}}.
\newblock


\bibitem[Koo et~al\mbox{.}(2023)]%
        {koo2023salad}
\bibfield{author}{\bibinfo{person}{Juil Koo}, \bibinfo{person}{Seungwoo Yoo}, \bibinfo{person}{Minh~Hieu Nguyen}, {and} \bibinfo{person}{Minhyuk Sung}.} \bibinfo{year}{2023}\natexlab{}.
\newblock \showarticletitle{Salad: Part-level latent diffusion for 3d shape generation and manipulation}. In \bibinfo{booktitle}{\emph{ICCV}}.
\newblock


\bibitem[Lan et~al\mbox{.}(2025)]%
        {lan2025ln3diff}
\bibfield{author}{\bibinfo{person}{Yushi Lan}, \bibinfo{person}{Fangzhou Hong}, \bibinfo{person}{Shuai Yang}, \bibinfo{person}{Shangchen Zhou}, \bibinfo{person}{Xuyi Meng}, \bibinfo{person}{Bo Dai}, \bibinfo{person}{Xingang Pan}, {and} \bibinfo{person}{Chen~Change Loy}.} \bibinfo{year}{2025}\natexlab{}.
\newblock \showarticletitle{Ln3diff: Scalable latent neural fields diffusion for speedy 3d generation}. In \bibinfo{booktitle}{\emph{ECCV}}.
\newblock


\bibitem[Li et~al\mbox{.}(2022)]%
        {li2022glip}
\bibfield{author}{\bibinfo{person}{Liunian~Harold Li}, \bibinfo{person}{Pengchuan Zhang}, \bibinfo{person}{Haotian Zhang}, \bibinfo{person}{Jianwei Yang}, \bibinfo{person}{Chunyuan Li}, \bibinfo{person}{Yiwu Zhong}, \bibinfo{person}{Lijuan Wang}, \bibinfo{person}{Lu Yuan}, \bibinfo{person}{Lei Zhang}, \bibinfo{person}{Jenq-Neng Hwang}, {et~al\mbox{.}}} \bibinfo{year}{2022}\natexlab{}.
\newblock \showarticletitle{Grounded language-image pre-training}. In \bibinfo{booktitle}{\emph{CVPR}}.
\newblock


\bibitem[Li et~al\mbox{.}(2023)]%
        {li2023diffusion}
\bibfield{author}{\bibinfo{person}{Muheng Li}, \bibinfo{person}{Yueqi Duan}, \bibinfo{person}{Jie Zhou}, {and} \bibinfo{person}{Jiwen Lu}.} \bibinfo{year}{2023}\natexlab{}.
\newblock \showarticletitle{Diffusion-sdf: Text-to-shape via voxelized diffusion}. In \bibinfo{booktitle}{\emph{CVPR}}.
\newblock


\bibitem[Li et~al\mbox{.}(2024b)]%
        {li2024pasta}
\bibfield{author}{\bibinfo{person}{Songlin Li}, \bibinfo{person}{Despoina Paschalidou}, {and} \bibinfo{person}{Leonidas Guibas}.} \bibinfo{year}{2024}\natexlab{b}.
\newblock \showarticletitle{PASTA: Controllable Part-Aware Shape Generation with Autoregressive Transformers}.
\newblock \bibinfo{journal}{\emph{arXiv preprint arXiv:2407.13677}} (\bibinfo{year}{2024}).
\newblock


\bibitem[Li et~al\mbox{.}(2024a)]%
        {li2024craftsman}
\bibfield{author}{\bibinfo{person}{Weiyu Li}, \bibinfo{person}{Jiarui Liu}, \bibinfo{person}{Rui Chen}, \bibinfo{person}{Yixun Liang}, \bibinfo{person}{Xuelin Chen}, \bibinfo{person}{Ping Tan}, {and} \bibinfo{person}{Xiaoxiao Long}.} \bibinfo{year}{2024}\natexlab{a}.
\newblock \showarticletitle{CraftsMan: High-fidelity Mesh Generation with 3D Native Generation and Interactive Geometry Refiner}.
\newblock \bibinfo{journal}{\emph{arXiv preprint arXiv:2405.14979}} (\bibinfo{year}{2024}).
\newblock


\bibitem[Li et~al\mbox{.}(2018)]%
        {li2018pointcnn}
\bibfield{author}{\bibinfo{person}{Yangyan Li}, \bibinfo{person}{Rui Bu}, \bibinfo{person}{Mingchao Sun}, \bibinfo{person}{Wei Wu}, \bibinfo{person}{Xinhan Di}, {and} \bibinfo{person}{Baoquan Chen}.} \bibinfo{year}{2018}\natexlab{}.
\newblock \showarticletitle{Pointcnn: Convolution on x-transformed points}. In \bibinfo{booktitle}{\emph{NeurIPS}}.
\newblock


\bibitem[Li et~al\mbox{.}(2025)]%
        {li2025triposg}
\bibfield{author}{\bibinfo{person}{Yangguang Li}, \bibinfo{person}{Zi-Xin Zou}, \bibinfo{person}{Zexiang Liu}, \bibinfo{person}{Dehu Wang}, \bibinfo{person}{Yuan Liang}, \bibinfo{person}{Zhipeng Yu}, \bibinfo{person}{Xingchao Liu}, \bibinfo{person}{Yuan-Chen Guo}, \bibinfo{person}{Ding Liang}, \bibinfo{person}{Wanli Ouyang}, {et~al\mbox{.}}} \bibinfo{year}{2025}\natexlab{}.
\newblock \showarticletitle{TripoSG: High-Fidelity 3D Shape Synthesis using Large-Scale Rectified Flow Models}.
\newblock \bibinfo{journal}{\emph{arXiv preprint arXiv:2502.06608}} (\bibinfo{year}{2025}).
\newblock


\bibitem[Lin et~al\mbox{.}(2025)]%
        {lin2025partcrafter}
\bibfield{author}{\bibinfo{person}{Yuchen Lin}, \bibinfo{person}{Chenguo Lin}, \bibinfo{person}{Panwang Pan}, \bibinfo{person}{Honglei Yan}, \bibinfo{person}{Yiqiang Feng}, \bibinfo{person}{Yadong Mu}, {and} \bibinfo{person}{Katerina Fragkiadaki}.} \bibinfo{year}{2025}\natexlab{}.
\newblock \showarticletitle{PartCrafter: Structured 3D Mesh Generation via Compositional Latent Diffusion Transformers}.
\newblock \bibinfo{journal}{\emph{arXiv preprint arXiv:2506.05573}} (\bibinfo{year}{2025}).
\newblock


\bibitem[Lipman et~al\mbox{.}(2024)]%
        {lipman2022flow}
\bibfield{author}{\bibinfo{person}{Yaron Lipman}, \bibinfo{person}{Ricky~TQ Chen}, \bibinfo{person}{Heli Ben-Hamu}, \bibinfo{person}{Maximilian Nickel}, {and} \bibinfo{person}{Matt Le}.} \bibinfo{year}{2024}\natexlab{}.
\newblock \showarticletitle{Flow matching for generative modeling}. In \bibinfo{booktitle}{\emph{NeurIPS}}.
\newblock


\bibitem[Liu et~al\mbox{.}(2024a)]%
        {liu2024part123}
\bibfield{author}{\bibinfo{person}{Anran Liu}, \bibinfo{person}{Cheng Lin}, \bibinfo{person}{Yuan Liu}, \bibinfo{person}{Xiaoxiao Long}, \bibinfo{person}{Zhiyang Dou}, \bibinfo{person}{Hao-Xiang Guo}, \bibinfo{person}{Ping Luo}, {and} \bibinfo{person}{Wenping Wang}.} \bibinfo{year}{2024}\natexlab{a}.
\newblock \showarticletitle{Part123: part-aware 3d reconstruction from a single-view image}. In \bibinfo{booktitle}{\emph{ACM SIGGRAPH}}.
\newblock


\bibitem[Liu et~al\mbox{.}(2025b)]%
        {liu2025freemesh}
\bibfield{author}{\bibinfo{person}{Jian Liu}, \bibinfo{person}{Haohan Weng}, \bibinfo{person}{Biwen Lei}, \bibinfo{person}{Xianghui Yang}, \bibinfo{person}{Zibo Zhao}, \bibinfo{person}{Zhuo Chen}, \bibinfo{person}{Song Guo}, \bibinfo{person}{Tao Han}, {and} \bibinfo{person}{Chunchao Guo}.} \bibinfo{year}{2025}\natexlab{b}.
\newblock \showarticletitle{FreeMesh: Boosting Mesh Generation with Coordinates Merging}.
\newblock \bibinfo{journal}{\emph{arXiv preprint arXiv:2505.13573}} (\bibinfo{year}{2025}).
\newblock


\bibitem[Liu et~al\mbox{.}(2024c)]%
        {liu2023one2345}
\bibfield{author}{\bibinfo{person}{Minghua Liu}, \bibinfo{person}{Ruoxi Shi}, \bibinfo{person}{Linghao Chen}, \bibinfo{person}{Zhuoyang Zhang}, \bibinfo{person}{Chao Xu}, \bibinfo{person}{Xinyue Wei}, \bibinfo{person}{Hansheng Chen}, \bibinfo{person}{Chong Zeng}, \bibinfo{person}{Jiayuan Gu}, {and} \bibinfo{person}{Hao Su}.} \bibinfo{year}{2024}\natexlab{c}.
\newblock \showarticletitle{One-2-3-45++: Fast Single Image to 3D Objects with Consistent Multi-View Generation and 3D Diffusion}. In \bibinfo{booktitle}{\emph{CVPR}}.
\newblock


\bibitem[Liu et~al\mbox{.}(2025a)]%
        {liu2025partfield}
\bibfield{author}{\bibinfo{person}{Minghua Liu}, \bibinfo{person}{Mikaela~Angelina Uy}, \bibinfo{person}{Donglai Xiang}, \bibinfo{person}{Hao Su}, \bibinfo{person}{Sanja Fidler}, \bibinfo{person}{Nicholas Sharp}, {and} \bibinfo{person}{Jun Gao}.} \bibinfo{year}{2025}\natexlab{a}.
\newblock \showarticletitle{PARTFIELD: Learning 3D Feature Fields for Part Segmentation and Beyond}.
\newblock \bibinfo{journal}{\emph{arXiv preprint arXiv:2504.11451}} (\bibinfo{year}{2025}).
\newblock


\bibitem[Liu et~al\mbox{.}(2023)]%
        {liu2023partslip}
\bibfield{author}{\bibinfo{person}{Minghua Liu}, \bibinfo{person}{Yinhao Zhu}, \bibinfo{person}{Hong Cai}, \bibinfo{person}{Shizhong Han}, \bibinfo{person}{Zhan Ling}, \bibinfo{person}{Fatih Porikli}, {and} \bibinfo{person}{Hao Su}.} \bibinfo{year}{2023}\natexlab{}.
\newblock \showarticletitle{Partslip: Low-shot part segmentation for 3d point clouds via pretrained image-language models}. In \bibinfo{booktitle}{\emph{CVPR}}.
\newblock


\bibitem[Liu et~al\mbox{.}(2024b)]%
        {liu2023syncdreamer}
\bibfield{author}{\bibinfo{person}{Yuan Liu}, \bibinfo{person}{Cheng Lin}, \bibinfo{person}{Zijiao Zeng}, \bibinfo{person}{Xiaoxiao Long}, \bibinfo{person}{Lingjie Liu}, \bibinfo{person}{Taku Komura}, {and} \bibinfo{person}{Wenping Wang}.} \bibinfo{year}{2024}\natexlab{b}.
\newblock \showarticletitle{SyncDreamer: Learning to Generate Multiview-consistent Images from a Single-view Image}. In \bibinfo{booktitle}{\emph{ICLR}}.
\newblock


\bibitem[Long et~al\mbox{.}(2024)]%
        {long2024wonder3d}
\bibfield{author}{\bibinfo{person}{Xiaoxiao Long}, \bibinfo{person}{Yuan-Chen Guo}, \bibinfo{person}{Cheng Lin}, \bibinfo{person}{Yuan Liu}, \bibinfo{person}{Zhiyang Dou}, \bibinfo{person}{Lingjie Liu}, \bibinfo{person}{Yuexin Ma}, \bibinfo{person}{Song-Hai Zhang}, \bibinfo{person}{Marc Habermann}, \bibinfo{person}{Christian Theobalt}, {et~al\mbox{.}}} \bibinfo{year}{2024}\natexlab{}.
\newblock \showarticletitle{Wonder3d: Single image to 3d using cross-domain diffusion}. In \bibinfo{booktitle}{\emph{CVPR}}.
\newblock


\bibitem[Mildenhall et~al\mbox{.}(2021)]%
        {mildenhall2021nerf}
\bibfield{author}{\bibinfo{person}{Ben Mildenhall}, \bibinfo{person}{Pratul~P Srinivasan}, \bibinfo{person}{Matthew Tancik}, \bibinfo{person}{Jonathan~T Barron}, \bibinfo{person}{Ravi Ramamoorthi}, {and} \bibinfo{person}{Ren Ng}.} \bibinfo{year}{2021}\natexlab{}.
\newblock \showarticletitle{Nerf: Representing scenes as neural radiance fields for view synthesis}.
\newblock \bibinfo{journal}{\emph{Commun. ACM}} \bibinfo{volume}{65}, \bibinfo{number}{1} (\bibinfo{year}{2021}), \bibinfo{pages}{99--106}.
\newblock


\bibitem[Oquab et~al\mbox{.}(2023)]%
        {oquab2023dinov2}
\bibfield{author}{\bibinfo{person}{Maxime Oquab}, \bibinfo{person}{Timoth{\'e}e Darcet}, \bibinfo{person}{Th{\'e}o Moutakanni}, \bibinfo{person}{Huy Vo}, \bibinfo{person}{Marc Szafraniec}, \bibinfo{person}{Vasil Khalidov}, \bibinfo{person}{Pierre Fernandez}, \bibinfo{person}{Daniel Haziza}, \bibinfo{person}{Francisco Massa}, \bibinfo{person}{Alaaeldin El-Nouby}, {et~al\mbox{.}}} \bibinfo{year}{2023}\natexlab{}.
\newblock \showarticletitle{Dinov2: Learning robust visual features without supervision}.
\newblock \bibinfo{journal}{\emph{arXiv preprint arXiv:2304.07193}} (\bibinfo{year}{2023}).
\newblock


\bibitem[Poole et~al\mbox{.}(2023)]%
        {poole2022dreamfusion}
\bibfield{author}{\bibinfo{person}{Ben Poole}, \bibinfo{person}{Ajay Jain}, \bibinfo{person}{Jonathan~T Barron}, {and} \bibinfo{person}{Ben Mildenhall}.} \bibinfo{year}{2023}\natexlab{}.
\newblock \showarticletitle{DreamFusion: Text-to-3D using 2D Diffusion}. In \bibinfo{booktitle}{\emph{ICLR}}.
\newblock


\bibitem[Pu et~al\mbox{.}(2025)]%
        {pu2025art}
\bibfield{author}{\bibinfo{person}{Yifan Pu}, \bibinfo{person}{Yiming Zhao}, \bibinfo{person}{Zhicong Tang}, \bibinfo{person}{Ruihong Yin}, \bibinfo{person}{Haoxing Ye}, \bibinfo{person}{Yuhui Yuan}, \bibinfo{person}{Dong Chen}, \bibinfo{person}{Jianmin Bao}, \bibinfo{person}{Sirui Zhang}, \bibinfo{person}{Yanbin Wang}, {et~al\mbox{.}}} \bibinfo{year}{2025}\natexlab{}.
\newblock \showarticletitle{Art: Anonymous region transformer for variable multi-layer transparent image generation}. In \bibinfo{booktitle}{\emph{CVPR}}.
\newblock


\bibitem[Qi et~al\mbox{.}(2017a)]%
        {qi2017pointnet}
\bibfield{author}{\bibinfo{person}{Charles~R Qi}, \bibinfo{person}{Hao Su}, \bibinfo{person}{Kaichun Mo}, {and} \bibinfo{person}{Leonidas~J Guibas}.} \bibinfo{year}{2017}\natexlab{a}.
\newblock \showarticletitle{Pointnet: Deep learning on point sets for 3d classification and segmentation}. In \bibinfo{booktitle}{\emph{CVPR}}.
\newblock


\bibitem[Qi et~al\mbox{.}(2017b)]%
        {qi2017pointnet++}
\bibfield{author}{\bibinfo{person}{Charles~Ruizhongtai Qi}, \bibinfo{person}{Li Yi}, \bibinfo{person}{Hao Su}, {and} \bibinfo{person}{Leonidas~J Guibas}.} \bibinfo{year}{2017}\natexlab{b}.
\newblock \showarticletitle{Pointnet++: Deep hierarchical feature learning on point sets in a metric space}. In \bibinfo{booktitle}{\emph{NeurIPS}}.
\newblock


\bibitem[Qi et~al\mbox{.}(2024)]%
        {qi2024tailor3d}
\bibfield{author}{\bibinfo{person}{Zhangyang Qi}, \bibinfo{person}{Yunhan Yang}, \bibinfo{person}{Mengchen Zhang}, \bibinfo{person}{Long Xing}, \bibinfo{person}{Xiaoyang Wu}, \bibinfo{person}{Tong Wu}, \bibinfo{person}{Dahua Lin}, \bibinfo{person}{Xihui Liu}, \bibinfo{person}{Jiaqi Wang}, {and} \bibinfo{person}{Hengshuang Zhao}.} \bibinfo{year}{2024}\natexlab{}.
\newblock \showarticletitle{Tailor3d: Customized 3d assets editing and generation with dual-side images}.
\newblock \bibinfo{journal}{\emph{arXiv preprint arXiv:2407.06191}} (\bibinfo{year}{2024}).
\newblock


\bibitem[Qian et~al\mbox{.}(2022)]%
        {qian2022pointnext}
\bibfield{author}{\bibinfo{person}{Guocheng Qian}, \bibinfo{person}{Yuchen Li}, \bibinfo{person}{Houwen Peng}, \bibinfo{person}{Jinjie Mai}, \bibinfo{person}{Hasan Hammoud}, \bibinfo{person}{Mohamed Elhoseiny}, {and} \bibinfo{person}{Bernard Ghanem}.} \bibinfo{year}{2022}\natexlab{}.
\newblock \showarticletitle{Pointnext: Revisiting pointnet++ with improved training and scaling strategies}. In \bibinfo{booktitle}{\emph{NeurIPS}}.
\newblock


\bibitem[Radford et~al\mbox{.}(2021)]%
        {radford2021clip}
\bibfield{author}{\bibinfo{person}{Alec Radford}, \bibinfo{person}{Jong~Wook Kim}, \bibinfo{person}{Chris Hallacy}, \bibinfo{person}{Aditya Ramesh}, \bibinfo{person}{Gabriel Goh}, \bibinfo{person}{Sandhini Agarwal}, \bibinfo{person}{Girish Sastry}, \bibinfo{person}{Amanda Askell}, \bibinfo{person}{Pamela Mishkin}, \bibinfo{person}{Jack Clark}, {et~al\mbox{.}}} \bibinfo{year}{2021}\natexlab{}.
\newblock \showarticletitle{Learning transferable visual models from natural language supervision}. In \bibinfo{booktitle}{\emph{ICML}}.
\newblock


\bibitem[Ravi et~al\mbox{.}(2024)]%
        {ravi2024sam}
\bibfield{author}{\bibinfo{person}{Nikhila Ravi}, \bibinfo{person}{Valentin Gabeur}, \bibinfo{person}{Yuan-Ting Hu}, \bibinfo{person}{Ronghang Hu}, \bibinfo{person}{Chaitanya Ryali}, \bibinfo{person}{Tengyu Ma}, \bibinfo{person}{Haitham Khedr}, \bibinfo{person}{Roman R{\"a}dle}, \bibinfo{person}{Chloe Rolland}, \bibinfo{person}{Laura Gustafson}, {et~al\mbox{.}}} \bibinfo{year}{2024}\natexlab{}.
\newblock \showarticletitle{Sam 2: Segment anything in images and videos}.
\newblock \bibinfo{journal}{\emph{arXiv preprint arXiv:2408.00714}} (\bibinfo{year}{2024}).
\newblock


\bibitem[Rombach et~al\mbox{.}(2022)]%
        {rombach2022high}
\bibfield{author}{\bibinfo{person}{Robin Rombach}, \bibinfo{person}{Andreas Blattmann}, \bibinfo{person}{Dominik Lorenz}, \bibinfo{person}{Patrick Esser}, {and} \bibinfo{person}{Bj{\"o}rn Ommer}.} \bibinfo{year}{2022}\natexlab{}.
\newblock \showarticletitle{High-resolution image synthesis with latent diffusion models}. In \bibinfo{booktitle}{\emph{CVPR}}.
\newblock


\bibitem[Shi et~al\mbox{.}(2023)]%
        {shi2023zero123plus}
\bibfield{author}{\bibinfo{person}{Ruoxi Shi}, \bibinfo{person}{Hansheng Chen}, \bibinfo{person}{Zhuoyang Zhang}, \bibinfo{person}{Minghua Liu}, \bibinfo{person}{Chao Xu}, \bibinfo{person}{Xinyue Wei}, \bibinfo{person}{Linghao Chen}, \bibinfo{person}{Chong Zeng}, {and} \bibinfo{person}{Hao Su}.} \bibinfo{year}{2023}\natexlab{}.
\newblock \showarticletitle{Zero123++: a single image to consistent multi-view diffusion base model}.
\newblock \bibinfo{journal}{\emph{arXiv preprint arXiv:2310.15110}} (\bibinfo{year}{2023}).
\newblock


\bibitem[Shim et~al\mbox{.}(2023)]%
        {shim2023diffusion}
\bibfield{author}{\bibinfo{person}{Jaehyeok Shim}, \bibinfo{person}{Changwoo Kang}, {and} \bibinfo{person}{Kyungdon Joo}.} \bibinfo{year}{2023}\natexlab{}.
\newblock \showarticletitle{Diffusion-based signed distance fields for 3d shape generation}. In \bibinfo{booktitle}{\emph{CVPR}}.
\newblock


\bibitem[Siddiqui et~al\mbox{.}(2024)]%
        {siddiqui2024meshgpt}
\bibfield{author}{\bibinfo{person}{Yawar Siddiqui}, \bibinfo{person}{Antonio Alliegro}, \bibinfo{person}{Alexey Artemov}, \bibinfo{person}{Tatiana Tommasi}, \bibinfo{person}{Daniele Sirigatti}, \bibinfo{person}{Vladislav Rosov}, \bibinfo{person}{Angela Dai}, {and} \bibinfo{person}{Matthias Nie{\ss}ner}.} \bibinfo{year}{2024}\natexlab{}.
\newblock \showarticletitle{Meshgpt: Generating triangle meshes with decoder-only transformers}. In \bibinfo{booktitle}{\emph{CVPR}}.
\newblock


\bibitem[Tang et~al\mbox{.}(2024)]%
        {tang2024segment}
\bibfield{author}{\bibinfo{person}{George Tang}, \bibinfo{person}{William Zhao}, \bibinfo{person}{Logan Ford}, \bibinfo{person}{David Benhaim}, {and} \bibinfo{person}{Paul Zhang}.} \bibinfo{year}{2024}\natexlab{}.
\newblock \showarticletitle{Segment Any Mesh: Zero-shot Mesh Part Segmentation via Lifting Segment Anything 2 to 3D}.
\newblock \bibinfo{journal}{\emph{arXiv:2408.13679}} (\bibinfo{year}{2024}).
\newblock


\bibitem[Tang et~al\mbox{.}(2025a)]%
        {tang2024edgerunner}
\bibfield{author}{\bibinfo{person}{Jiaxiang Tang}, \bibinfo{person}{Zhaoshuo Li}, \bibinfo{person}{Zekun Hao}, \bibinfo{person}{Xian Liu}, \bibinfo{person}{Gang Zeng}, \bibinfo{person}{Ming-Yu Liu}, {and} \bibinfo{person}{Qinsheng Zhang}.} \bibinfo{year}{2025}\natexlab{a}.
\newblock \showarticletitle{Edgerunner: Auto-regressive auto-encoder for artistic mesh generation}. In \bibinfo{booktitle}{\emph{ICLR}}.
\newblock


\bibitem[Tang et~al\mbox{.}(2025b)]%
        {tang2025efficient}
\bibfield{author}{\bibinfo{person}{Jiaxiang Tang}, \bibinfo{person}{Ruijie Lu}, \bibinfo{person}{Zhaoshuo Li}, \bibinfo{person}{Zekun Hao}, \bibinfo{person}{Xuan Li}, \bibinfo{person}{Fangyin Wei}, \bibinfo{person}{Shuran Song}, \bibinfo{person}{Gang Zeng}, \bibinfo{person}{Ming-Yu Liu}, {and} \bibinfo{person}{Tsung-Yi Lin}.} \bibinfo{year}{2025}\natexlab{b}.
\newblock \showarticletitle{Efficient Part-level 3D Object Generation via Dual Volume Packing}.
\newblock \bibinfo{journal}{\emph{arXiv preprint arXiv:2506.09980}} (\bibinfo{year}{2025}).
\newblock


\bibitem[Thai et~al\mbox{.}(2024)]%
        {thai20243x2}
\bibfield{author}{\bibinfo{person}{Anh Thai}, \bibinfo{person}{Weiyao Wang}, \bibinfo{person}{Hao Tang}, \bibinfo{person}{Stefan Stojanov}, \bibinfo{person}{Matt Feiszli}, {and} \bibinfo{person}{James~M Rehg}.} \bibinfo{year}{2024}\natexlab{}.
\newblock \showarticletitle{3x2: 3D Object Part Segmentation by 2D Semantic Correspondences}. In \bibinfo{booktitle}{\emph{ECCV}}.
\newblock


\bibitem[Wang et~al\mbox{.}(2025)]%
        {wang2025nautilus}
\bibfield{author}{\bibinfo{person}{Yuxuan Wang}, \bibinfo{person}{Xuanyu Yi}, \bibinfo{person}{Haohan Weng}, \bibinfo{person}{Qingshan Xu}, \bibinfo{person}{Xiaokang Wei}, \bibinfo{person}{Xianghui Yang}, \bibinfo{person}{Chunchao Guo}, \bibinfo{person}{Long Chen}, {and} \bibinfo{person}{Hanwang Zhang}.} \bibinfo{year}{2025}\natexlab{}.
\newblock \showarticletitle{Nautilus: Locality-aware Autoencoder for Scalable Mesh Generation}.
\newblock \bibinfo{journal}{\emph{arXiv preprint arXiv:2501.14317}} (\bibinfo{year}{2025}).
\newblock


\bibitem[Weng et~al\mbox{.}(2024)]%
        {weng2024scaling}
\bibfield{author}{\bibinfo{person}{Haohan Weng}, \bibinfo{person}{Zibo Zhao}, \bibinfo{person}{Biwen Lei}, \bibinfo{person}{Xianghui Yang}, \bibinfo{person}{Jian Liu}, \bibinfo{person}{Zeqiang Lai}, \bibinfo{person}{Zhuo Chen}, \bibinfo{person}{Yuhong Liu}, \bibinfo{person}{Jie Jiang}, \bibinfo{person}{Chunchao Guo}, {et~al\mbox{.}}} \bibinfo{year}{2024}\natexlab{}.
\newblock \showarticletitle{Scaling mesh generation via compressive tokenization}.
\newblock \bibinfo{journal}{\emph{arXiv preprint arXiv:2411.07025}} (\bibinfo{year}{2024}).
\newblock


\bibitem[Wu et~al\mbox{.}(2024)]%
        {wu2024direct3d}
\bibfield{author}{\bibinfo{person}{Shuang Wu}, \bibinfo{person}{Youtian Lin}, \bibinfo{person}{Feihu Zhang}, \bibinfo{person}{Yifei Zeng}, \bibinfo{person}{Jingxi Xu}, \bibinfo{person}{Philip Torr}, \bibinfo{person}{Xun Cao}, {and} \bibinfo{person}{Yao Yao}.} \bibinfo{year}{2024}\natexlab{}.
\newblock \showarticletitle{Direct3D: Scalable Image-to-3D Generation via 3D Latent Diffusion Transformer}. In \bibinfo{booktitle}{\emph{NeurIPS}}.
\newblock


\bibitem[Xiang et~al\mbox{.}(2024)]%
        {xiang2024structured}
\bibfield{author}{\bibinfo{person}{Jianfeng Xiang}, \bibinfo{person}{Zelong Lv}, \bibinfo{person}{Sicheng Xu}, \bibinfo{person}{Yu Deng}, \bibinfo{person}{Ruicheng Wang}, \bibinfo{person}{Bowen Zhang}, \bibinfo{person}{Dong Chen}, \bibinfo{person}{Xin Tong}, {and} \bibinfo{person}{Jiaolong Yang}.} \bibinfo{year}{2024}\natexlab{}.
\newblock \showarticletitle{Structured 3d latents for scalable and versatile 3d generation}.
\newblock \bibinfo{journal}{\emph{arXiv preprint arXiv:2412.01506}} (\bibinfo{year}{2024}).
\newblock


\bibitem[Xu et~al\mbox{.}(2024)]%
        {xu2024instantmesh}
\bibfield{author}{\bibinfo{person}{Jiale Xu}, \bibinfo{person}{Weihao Cheng}, \bibinfo{person}{Yiming Gao}, \bibinfo{person}{Xintao Wang}, \bibinfo{person}{Shenghua Gao}, {and} \bibinfo{person}{Ying Shan}.} \bibinfo{year}{2024}\natexlab{}.
\newblock \showarticletitle{Instantmesh: Efficient 3d mesh generation from a single image with sparse-view large reconstruction models}.
\newblock \bibinfo{journal}{\emph{arXiv preprint arXiv:2404.07191}} (\bibinfo{year}{2024}).
\newblock


\bibitem[Xue et~al\mbox{.}(2023)]%
        {xue2023zerops}
\bibfield{author}{\bibinfo{person}{Yuheng Xue}, \bibinfo{person}{Nenglun Chen}, \bibinfo{person}{Jun Liu}, {and} \bibinfo{person}{Wenyun Sun}.} \bibinfo{year}{2023}\natexlab{}.
\newblock \showarticletitle{ZeroPS: High-quality Cross-modal Knowledge Transfer for Zero-Shot 3D Part Segmentation}.
\newblock \bibinfo{journal}{\emph{arXiv:2311.14262}} (\bibinfo{year}{2023}).
\newblock


\bibitem[Yan et~al\mbox{.}(2024)]%
        {yan2024phycage}
\bibfield{author}{\bibinfo{person}{Han Yan}, \bibinfo{person}{Mingrui Zhang}, \bibinfo{person}{Yang Li}, \bibinfo{person}{Chao Ma}, {and} \bibinfo{person}{Pan Ji}.} \bibinfo{year}{2024}\natexlab{}.
\newblock \showarticletitle{PhyCAGE: Physically Plausible Compositional 3D Asset Generation from a Single Image}.
\newblock \bibinfo{journal}{\emph{arXiv preprint arXiv:2411.18548}} (\bibinfo{year}{2024}).
\newblock


\bibitem[Yang et~al\mbox{.}(2025a)]%
        {yang2025dreamcomposer++}
\bibfield{author}{\bibinfo{person}{Yunhan Yang}, \bibinfo{person}{Shuo Chen}, \bibinfo{person}{Yukun Huang}, \bibinfo{person}{Xiaoyang Wu}, \bibinfo{person}{Yuan-Chen Guo}, \bibinfo{person}{Edmund~Y Lam}, \bibinfo{person}{Hengshuang Zhao}, \bibinfo{person}{Tong He}, {and} \bibinfo{person}{Xihui Liu}.} \bibinfo{year}{2025}\natexlab{a}.
\newblock \showarticletitle{DreamComposer++: Empowering Diffusion Models with Multi-View Conditions for 3D Content Generation}.
\newblock \bibinfo{journal}{\emph{IEEE Transactions on Pattern Analysis and Machine Intelligence}} (\bibinfo{year}{2025}).
\newblock


\bibitem[Yang et~al\mbox{.}(2025b)]%
        {yang2025holopart}
\bibfield{author}{\bibinfo{person}{Yunhan Yang}, \bibinfo{person}{Yuan-Chen Guo}, \bibinfo{person}{Yukun Huang}, \bibinfo{person}{Zi-Xin Zou}, \bibinfo{person}{Zhipeng Yu}, \bibinfo{person}{Yangguang Li}, \bibinfo{person}{Yan-Pei Cao}, {and} \bibinfo{person}{Xihui Liu}.} \bibinfo{year}{2025}\natexlab{b}.
\newblock \showarticletitle{HoloPart: Generative 3D Part Amodal Segmentation}.
\newblock \bibinfo{journal}{\emph{arXiv preprint arXiv:2504.07943}} (\bibinfo{year}{2025}).
\newblock


\bibitem[Yang et~al\mbox{.}(2024a)]%
        {yang2024sampart3dsegment3dobjects}
\bibfield{author}{\bibinfo{person}{Yunhan Yang}, \bibinfo{person}{Yukun Huang}, \bibinfo{person}{Yuan-Chen Guo}, \bibinfo{person}{Liangjun Lu}, \bibinfo{person}{Xiaoyang Wu}, \bibinfo{person}{Edmund~Y Lam}, \bibinfo{person}{Yan-Pei Cao}, {and} \bibinfo{person}{Xihui Liu}.} \bibinfo{year}{2024}\natexlab{a}.
\newblock \showarticletitle{Sampart3d: Segment any part in 3d objects}.
\newblock \bibinfo{journal}{\emph{arXiv preprint arXiv:2411.07184}} (\bibinfo{year}{2024}).
\newblock


\bibitem[Yang et~al\mbox{.}(2024b)]%
        {yang2024dreamcomposer}
\bibfield{author}{\bibinfo{person}{Yunhan Yang}, \bibinfo{person}{Yukun Huang}, \bibinfo{person}{Xiaoyang Wu}, \bibinfo{person}{Yuan-Chen Guo}, \bibinfo{person}{Song-Hai Zhang}, \bibinfo{person}{Hengshuang Zhao}, \bibinfo{person}{Tong He}, {and} \bibinfo{person}{Xihui Liu}.} \bibinfo{year}{2024}\natexlab{b}.
\newblock \showarticletitle{{DreamComposer: Controllable 3D Object Generation via Multi-View Conditions}}. In \bibinfo{booktitle}{\emph{CVPR}}.
\newblock


\bibitem[Yang et~al\mbox{.}(2023)]%
        {yang2023sam3d}
\bibfield{author}{\bibinfo{person}{Yunhan Yang}, \bibinfo{person}{Xiaoyang Wu}, \bibinfo{person}{Tong He}, \bibinfo{person}{Hengshuang Zhao}, {and} \bibinfo{person}{Xihui Liu}.} \bibinfo{year}{2023}\natexlab{}.
\newblock \showarticletitle{Sam3d: Segment anything in 3d scenes}.
\newblock \bibinfo{journal}{\emph{arXiv preprint arXiv:2306.03908}} (\bibinfo{year}{2023}).
\newblock


\bibitem[Zhang et~al\mbox{.}(2023)]%
        {zhang20233dshape2vecset}
\bibfield{author}{\bibinfo{person}{Biao Zhang}, \bibinfo{person}{Jiapeng Tang}, \bibinfo{person}{Matthias Niessner}, {and} \bibinfo{person}{Peter Wonka}.} \bibinfo{year}{2023}\natexlab{}.
\newblock \showarticletitle{3dshape2vecset: A 3d shape representation for neural fields and generative diffusion models}.
\newblock \bibinfo{journal}{\emph{ACM Transactions On Graphics (TOG)}} \bibinfo{volume}{42}, \bibinfo{number}{4} (\bibinfo{year}{2023}), \bibinfo{pages}{1--16}.
\newblock


\bibitem[Zhang et~al\mbox{.}(2025)]%
        {zhang2025compress3d}
\bibfield{author}{\bibinfo{person}{Bowen Zhang}, \bibinfo{person}{Tianyu Yang}, \bibinfo{person}{Yu Li}, \bibinfo{person}{Lei Zhang}, {and} \bibinfo{person}{Xi Zhao}.} \bibinfo{year}{2025}\natexlab{}.
\newblock \showarticletitle{Compress3D: a compressed latent space for 3D generation from a single image}. In \bibinfo{booktitle}{\emph{ECCV}}.
\newblock


\bibitem[Zhang et~al\mbox{.}(2024a)]%
        {zhang2024compass}
\bibfield{author}{\bibinfo{person}{Gaoyang Zhang}, \bibinfo{person}{Bingtao Fu}, \bibinfo{person}{Qingnan Fan}, \bibinfo{person}{Qi Zhang}, \bibinfo{person}{Runxing Liu}, \bibinfo{person}{Hong Gu}, \bibinfo{person}{Huaqi Zhang}, {and} \bibinfo{person}{Xinguo Liu}.} \bibinfo{year}{2024}\natexlab{a}.
\newblock \showarticletitle{CoMPaSS: Enhancing Spatial Understanding in Text-to-Image Diffusion Models}.
\newblock \bibinfo{journal}{\emph{arXiv preprint arXiv:2412.13195}} (\bibinfo{year}{2024}).
\newblock


\bibitem[Zhang and Agrawala(2024)]%
        {zhang2024transparent}
\bibfield{author}{\bibinfo{person}{Lvmin Zhang} {and} \bibinfo{person}{Maneesh Agrawala}.} \bibinfo{year}{2024}\natexlab{}.
\newblock \showarticletitle{Transparent Image Layer Diffusion using Latent Transparency}.
\newblock \bibinfo{journal}{\emph{ACM Transactions on Graphics (TOG)}} \bibinfo{volume}{43}, \bibinfo{number}{4} (\bibinfo{year}{2024}), \bibinfo{pages}{1--15}.
\newblock


\bibitem[Zhang et~al\mbox{.}(2024b)]%
        {zhang2024clay}
\bibfield{author}{\bibinfo{person}{Longwen Zhang}, \bibinfo{person}{Ziyu Wang}, \bibinfo{person}{Qixuan Zhang}, \bibinfo{person}{Qiwei Qiu}, \bibinfo{person}{Anqi Pang}, \bibinfo{person}{Haoran Jiang}, \bibinfo{person}{Wei Yang}, \bibinfo{person}{Lan Xu}, {and} \bibinfo{person}{Jingyi Yu}.} \bibinfo{year}{2024}\natexlab{b}.
\newblock \showarticletitle{CLAY: A Controllable Large-scale Generative Model for Creating High-quality 3D Assets}.
\newblock \bibinfo{journal}{\emph{ACM Transactions on Graphics (TOG)}} \bibinfo{volume}{43}, \bibinfo{number}{4} (\bibinfo{year}{2024}), \bibinfo{pages}{1--20}.
\newblock


\bibitem[Zhang et~al\mbox{.}(2022)]%
        {zhang2022opt}
\bibfield{author}{\bibinfo{person}{Susan Zhang}, \bibinfo{person}{Stephen Roller}, \bibinfo{person}{Naman Goyal}, \bibinfo{person}{Mikel Artetxe}, \bibinfo{person}{Moya Chen}, \bibinfo{person}{Shuohui Chen}, \bibinfo{person}{Christopher Dewan}, \bibinfo{person}{Mona Diab}, \bibinfo{person}{Xian Li}, \bibinfo{person}{Xi~Victoria Lin}, {et~al\mbox{.}}} \bibinfo{year}{2022}\natexlab{}.
\newblock \showarticletitle{Opt: Open pre-trained transformer language models}.
\newblock \bibinfo{journal}{\emph{arXiv preprint arXiv:2205.01068}} (\bibinfo{year}{2022}).
\newblock


\bibitem[Zhao et~al\mbox{.}(2021)]%
        {zhao2021point}
\bibfield{author}{\bibinfo{person}{Hengshuang Zhao}, \bibinfo{person}{Li Jiang}, \bibinfo{person}{Jiaya Jia}, \bibinfo{person}{Philip~HS Torr}, {and} \bibinfo{person}{Vladlen Koltun}.} \bibinfo{year}{2021}\natexlab{}.
\newblock \showarticletitle{Point transformer}. In \bibinfo{booktitle}{\emph{ICCV}}.
\newblock


\bibitem[Zhao et~al\mbox{.}(2025)]%
        {zhao2025deepmesh}
\bibfield{author}{\bibinfo{person}{Ruowen Zhao}, \bibinfo{person}{Junliang Ye}, \bibinfo{person}{Zhengyi Wang}, \bibinfo{person}{Guangce Liu}, \bibinfo{person}{Yiwen Chen}, \bibinfo{person}{Yikai Wang}, {and} \bibinfo{person}{Jun Zhu}.} \bibinfo{year}{2025}\natexlab{}.
\newblock \showarticletitle{DeepMesh: Auto-Regressive Artist-mesh Creation with Reinforcement Learning}.
\newblock \bibinfo{journal}{\emph{arXiv preprint arXiv:2503.15265}} (\bibinfo{year}{2025}).
\newblock


\bibitem[Zhao et~al\mbox{.}(2024)]%
        {zhao2024michelangelo}
\bibfield{author}{\bibinfo{person}{Zibo Zhao}, \bibinfo{person}{Wen Liu}, \bibinfo{person}{Xin Chen}, \bibinfo{person}{Xianfang Zeng}, \bibinfo{person}{Rui Wang}, \bibinfo{person}{Pei Cheng}, \bibinfo{person}{Bin Fu}, \bibinfo{person}{Tao Chen}, \bibinfo{person}{Gang Yu}, {and} \bibinfo{person}{Shenghua Gao}.} \bibinfo{year}{2024}\natexlab{}.
\newblock \showarticletitle{Michelangelo: Conditional 3d shape generation based on shape-image-text aligned latent representation}. In \bibinfo{booktitle}{\emph{NeurIPS}}.
\newblock


\bibitem[Zhong et~al\mbox{.}(2024)]%
        {zhong2024meshsegmenter}
\bibfield{author}{\bibinfo{person}{Ziming Zhong}, \bibinfo{person}{Yanyu Xu}, \bibinfo{person}{Jing Li}, \bibinfo{person}{Jiale Xu}, \bibinfo{person}{Zhengxin Li}, \bibinfo{person}{Chaohui Yu}, {and} \bibinfo{person}{Shenghua Gao}.} \bibinfo{year}{2024}\natexlab{}.
\newblock \showarticletitle{MeshSegmenter: Zero-Shot Mesh Semantic Segmentation via Texture Synthesis}. In \bibinfo{booktitle}{\emph{ECCV}}.
\newblock


\bibitem[Zou et~al\mbox{.}(2024)]%
        {zou2024triplane}
\bibfield{author}{\bibinfo{person}{Zi-Xin Zou}, \bibinfo{person}{Zhipeng Yu}, \bibinfo{person}{Yuan-Chen Guo}, \bibinfo{person}{Yangguang Li}, \bibinfo{person}{Ding Liang}, \bibinfo{person}{Yan-Pei Cao}, {and} \bibinfo{person}{Song-Hai Zhang}.} \bibinfo{year}{2024}\natexlab{}.
\newblock \showarticletitle{Triplane meets gaussian splatting: Fast and generalizable single-view 3d reconstruction with transformers}. In \bibinfo{booktitle}{\emph{CVPR}}.
\newblock


\end{thebibliography}

\end{document}